\documentclass[10pt,twocolumn,letterpaper]{article}

 \usepackage{cvpr}              

\usepackage{todonotes}
\usepackage{pifont}
\usepackage{tikz}
\usepackage{adjustbox}
\usepackage{multirow}
\usepackage{booktabs,array}
\usetikzlibrary{shapes,arrows,positioning}

\usepackage{xcolor}
\usepackage{graphicx}

\usepackage{rotating}
\usepackage{tabularx}
\usepackage{amssymb}
\usepackage{pifont}
\newcommand{\xmark}{\ding{55}}%
\usepackage{booktabs}
\usepackage{placeins}

\newcommand{\greencheck}{{\color{green}\checkmark}}
\newcommand{\redxmark}{{\color{red}\xmark}}

\usepackage[acronym]{glossaries}
\newacronym{ALS}{ALS}{airborne laser scanning}
\newacronym{MLS}{MLS}{mobile laser scanning}
\newacronym{LoD}{LoD}{level of detail}
\newacronym{LoDs}{LoDs}{level of details}
\newacronym{OGC}{OGC}{Open Geospatial Consortium}
\newacronym{GML}{GML}{Geography Markup Language}
\newacronym{ASAM}{ASAM}{Association for Standardization of Automation and Measuring Systems}
\newacronym{TLS}{TLS}{terrestrial laser scanning}
\newacronym{UAV}{UAV}{unmanned aerial vehicle}
\newacronym{HD}{HD}{high definition}
\newacronym{RANSAC}{RANSAC}{RANdom SAmple Consensus}
\newacronym{ROI}{ROI}{region of interest}
\newacronym{DEM}{DEM}{digital elevation model}
\newacronym{ICP}{ICP}{iterative closest point}
\newacronym{NLOS}{NLOS}{non-line-of-sight}
\newacronym{SfM}{SfM}{structure from motion}
\newacronym{FME}{FME}{Feature Manipulation Engine}
\newacronym{OSM}{OSM}{OpenStreetMap} 
\newacronym{RMSE}{RMSE}{root mean square error}
\newacronym{CPT}{CPT}{conditional probability table}
\newacronym{DST}{DST}{Dempster–Shafer theory}
\newacronym{BN}{BayNet}{Bayesian network}
\newacronym{GIS}{GIS}{Geographic Information System}
\newacronym{PPD}{PPD}{posterior probability distribution}
\newacronym{CI}{CI}{confidence interval}
\newacronym{IFC}{IFC}{Industry Foundation Classes}
\newacronym{CRS}{CRS}{coordinate reference system}
\newacronym{LoFG}{LoFG}{Level of Facade Generalization}
\newacronym{RTK}{RTK}{real-time kinematic}
\newacronym{UTM}{UTM}{Universal Transverse Mercator}
\newacronym{IoU}{IoU}{Intersection-over-Union}
\newacronym{mIoU}{mIoU}{mean Intersection-over-Union}
\newacronym{mAcc}{mAcc}{mean Accuracy}
\newacronym{OA}{OA}{Overall Accuracy}

\usepackage{colortbl}

\usepackage{xcolor}
\definecolor{semanticLabelRoadSurface}{rgb}{0.3843137254901961,0.4549019607843137,0.5568627450980392}
\definecolor{semanticLabelGroundSurface}{rgb}{0.11372549019607843,0.1607843137254902,0.23921568627450981}
\definecolor{semanticLabelCityFurniture}{rgb}{1.0,0.5372549019607843,0.01568627450980392}
\definecolor{semanticLabelVehicle}{rgb}{1.0,0.39215686274509803,0.403921568627451}
\definecolor{semanticLabelPedestrian}{rgb}{0.9647058823529412,0.2,0.6039215686274509}
\definecolor{semanticLabelWallSurface}{rgb}{0.996078431372549,0.9764705882352941,0.7607843137254902}
\definecolor{semanticLabelRoofSurface}{rgb}{0.9058823529411765,0.0,0.043137254901960784}
\definecolor{semanticLabelDoor}{rgb}{0.6509803921568628,0.37254901960784315,0.0}
\definecolor{semanticLabelWindow}{rgb}{0.08235294117647059,0.36470588235294116,0.9882352941176471}
\definecolor{semanticLabelBuildingInstallation}{rgb}{0.7607843137254902,0.47843137254901963,1.0}
\definecolor{semanticLabelSolitaryVegetationObject}{rgb}{0.0,0.5098039215686274,0.21176470588235294}
\definecolor{semanticLabelNoise}{rgb}{0.6509803921568628,0.6274509803921569,0.6078431372549019}

\usepackage[dvipsnames]{xcolor}

\definecolor{cvprblue}{rgb}{0.21,0.49,0.74}
\usepackage[pagebackref,breaklinks,colorlinks,citecolor=cvprblue]{hyperref}

\def\paperID{302} 
\def\confName{3DV\xspace}
\def\confYear{2026\xspace}

\title{TrueCity: Real and Simulated Urban Data for Cross-Domain 3D Scene Understanding}

\author{Duc Nguyen\textsuperscript{*1 }, Yan-Ling Lai\textsuperscript{*1 }, Qilin Zhang\textsuperscript{1 }, Prabin Gyawali\textsuperscript{1 }, Benedikt Schwab\textsuperscript{1 }, \\ Olaf Wysocki\textsuperscript{1,2 }, Thomas H. Kolbe\textsuperscript{1 }\\ \\
\textsuperscript{1 }Technical University of Munich, \textsuperscript{2 } CV4DT, University of Cambridge \\\\
{\tt\small (duc.nguyen, ... thomas.kolbe)@tum.de} ;
{\tt\small okw24@cam.ac.uk}\\
{\tt\small * equal contribution}\\
}

\begin{document}
\maketitle
\begin{abstract}
3D semantic scene understanding remains a long-standing challenge in the 3D computer vision community.
One of the key issues pertains to limited real-world annotated data to facilitate generalizable models.
The common practice to tackle this issue is to simulate new data.  
Although synthetic datasets offer scalability and perfect labels, their designer-crafted scenes fail to capture real-world complexity and sensor noise, resulting in a synthetic-to-real domain gap. 
Moreover, no benchmark provides synchronized real and simulated point clouds for segmentation-oriented domain shift analysis.
We introduce TrueCity, the first urban semantic segmentation benchmark with cm-accurate annotated real-world point clouds, semantic 3D city models, and annotated simulated point clouds representing the same city.
TrueCity proposes segmentation classes aligned with international 3D city modeling standards, enabling consistent evaluation of synthetic-to-real gap. 
Our extensive experiments on common baselines quantify domain shift and highlight strategies for exploiting synthetic data to enhance real-world 3D scene understanding.
We are convinced that the TrueCity dataset will foster further development of sim-to-real gap quantification and enable generalizable data-driven models.
The data, code, and 3D models are available online: \url{https://tum-gis.github.io/TrueCity/}.
\end{abstract}
    
\section{Introduction}
\label{sec:intro}

\begin{figure}[htb]
    \centering
    \includegraphics[width=0.95\linewidth]{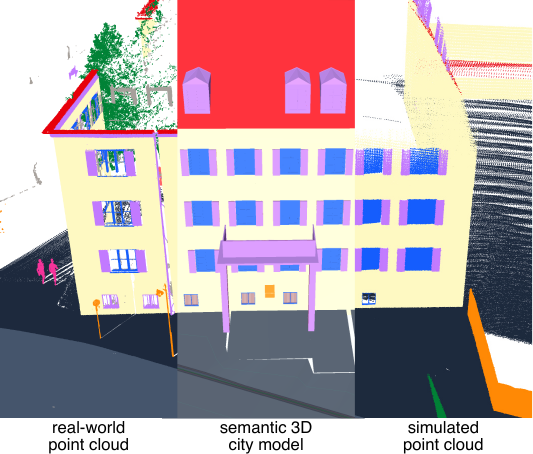}
    \caption{TrueCity introduces real-world annotated point clouds, a semantic 3D city model, and 3D-model simulated point clouds of the same location, enabling coherent evaluation of the sim-to-real domain gap in 3D scene understanding.} 
    \label{fig:teaser}
\end{figure}
Semantic segmentation of point clouds remains a critical yet unsolved task in 3D computer vision. 
One of the main impediments in this area is the limited availability of high-quality, semantically annotated real-world 3D data. 
To address this issue, many researchers have turned to synthetic data generation by simulating sensory inputs in virtual environments \cite{carla_paris,spiegel2021,huch2023quantifying,dosovitskiy2017carla,shah2017airsim}. 
These synthetic datasets offer an appealing alternative due to their scalability, cost-effectiveness, and precise ground truth annotations.

However, a significant drawback of using simulated data lies in the large domain shift between synthetic and real-world point clouds.
This is especially apparent when dealing with urban scenes which are highly challenging scenarios owing to changing scanning distances, various object material properties, and many dynamic events.
Simulated environments often rely on fictitious and designer-crafted 3D scenes that fail to capture the full complexity, variability, and noise present in real-world settings \cite{dosovitskiy2017carla,shah2017airsim,spiegel2021}.
Another limitation is the lack of benchmark datasets comprising synchronized real-world 3D point clouds and 3D environment paired with their simulated counterparts. 
Lack of such data hampers comprehensive quantification of the domain shift between synthetic and real domains. 
This research gap not only impedes the development of robust segmentation models but also limits our understanding of how synthetic data can best be leveraged to improve real-world performance.

Another general challenge in developing urban segmentation methods is the high heterogeneity and definition of urban classes.
This variability often leads to misinterpretation of object characteristics and hinders the creation of large-scale, high-diversity datasets.
In particular, there is often a taxonomic and geometric mismatch between the segmented urban elements and the internationally recognized and standardized modeling classes.
The issue is frequently overlooked, yet it poses a significant obstacle to analyze per class sim-to-real shift, especially apparent in ray-penetrable glass objects which simulation fails to capture \cite{wysocki2023scan2lod3,carla_paris}.

To address these challenges, our TrueCity contributes in:
\begin{itemize}
    \item Proposing novel urban semantic segmentation classes derived from international modeling standards;
    \item Realizing the new class design on the introduced TrueCity benchmark dataset comprising real-world cm-accurate labeled point clouds, its derived semantic 3D city model structured according to the CityGML 2.0 standard, and simulated point clouds representing the same city (\cref{fig:teaser});
    \item Analyzing domain shift between synthetic and real point clouds on the segmentation task enabled by data synchronization. 
\end{itemize}


\section{Related Work}
\label{sec:rw}
\subsection{Domain Shift in Point Cloud Segmentation }
\label{sec:SemSeg}
Inherently, machine and deep learning methods require a large amount of training data, which is expensive to collect.
Especially in the context of semantic segmentation of 3D point clouds, the scarcity of well-annotated data for each sensor type has become a notable challenge \citep{are_we_hungry,mind_the_gap,xiao2022transfer}. 
The immediate application of a model trained on one sensor type to another is often infeasible owing to the different laser scanning patterns, point distribution, and point cloud size.
For instance, \citet{SegTrans} show semantic segmentation decreases its performance significantly when trained on \gls{TLS} and inferred on \gls{MLS} point cloud, i.e., 76.5\% to 32.0\% in their experiments.
Recent research endeavors have shifted towards integrating domain adaptation techniques accounting for the sensor differences and showing promising results \citep{xiao2022transfer}, e.g., improving segmentation accuracy by up to 13\% \cite{SegTrans}.
Yet, the performance is still insufficiently robust, showing large variations across datasets and reaching only around 56\% IoU on challenging datasets \cite{wang2024unsupervised}.

Furthermore, unlike in 2D image domain \citep{ros2016synthia}, there is a lack of large real-world point clouds to train a generic classifier that can be adapted for multiple scenarios \citep{wysocki2025zaha}.
Consequently, recent years have witnessed a surge in methods investigating the adoption of simulated point clouds complementing real-world scenarios \citep{sim2real,xiao2022transfer,carla_paris}.
Current simulators often represent fictitious scenes designed manually \cite{dosovitskiy2017carla,shah2017airsim,winiwarterVirtualLaserScanning2022}.
This fact poses yet another challenge of not only accounting for the sensor type differences but also the domain gap between the simulated and real-world data. 
As reflected by experiments conducted by \citet{xiao2022transfer}, the performance for combining simulated and real data can oscillate in the range of 55-65\% accuracy depending on the scenario and benchmark dataset, underscoring the importance of further domain-gap investigations.

Although recently researchers have shown that weakly-supervised  \citep{lin2022weakly,wang2024survey} and self-supervised \citep{zeng2024self,zhang2021self} methods show promising results, there are still necessitating validation sets and their performance can be limited, e.g., can reach only up to 60\% when deployed for object detection, as shown in the review of \citet{zeng2024self}.

Worth noting is also the parallel submission to a journal \cite{ourPrevious}. 
There, the focus lies on providing the workflow to generate simulated data and provide deterministic and stochastic tools to validate the simulation.
Also, unlike in this TrueCity submission, no synchronized dataset is introduced; TrueCity further harmonizes the semantic classes with the international standards; and importantly TrueCity provides revised test, validation, and training split without mixing real with synthetic point clouds in local neighborhoods, thus enabling consistent domain shift evaluation in separate spatial regions.

\subsection{Urban Point Cloud Segmentation Datasets}
\label{sec:datasets}
A wide range of point cloud semantic segmentation benchmarks have been introduced in recent years (\cref{tab:bigTable}).
However, the availability of such datasets remains much smaller that of 2D image segmentation benchmarks, both in terms of the number of datasets and the volume of annotated instances. 
Street-level urban point cloud segmentation datasets represent particularly challenging scenarios, where even transformer-based methods show limited performance, e.g., Point Transformer achieves only around 42\% IoU on the facade segmentation task in the ZAHA dataset \cite{wysocki2025zaha}.

Despite progress, none of the current benchmarks provide synchronized real and simulated point clouds captured at the same geographic location (except aerial data which is out of scope for this publication \cite{chen2022stpls3d}).
Notable datasets such as DELIVER~\cite{liao2025benchmarking} and Paris-CARLA-3D~\cite{carla_paris} include simulation data, but the synthetic scenes are not representing the corresponding real sites. 
In Paris-CARLA-3D, Paris scans are paired with CARLA-generated fictitious towns, and the underlying 3D real city assets are unavailable, precluding scene-level re-simulation and making systematic analysis of domain shifts, and thus a fair estimate of the domain gap difficult.


\begin{table*}[!htb]
    \captionsetup{size=footnotesize}
    \centering
    \begin{adjustbox}{width=1\textwidth, center}
        \begin{tabular}{lccccccccc}
            \toprule
            Name  & Year &   Sensor   &   Acquisition?   &  \# Classes   & Real city?  & 3D models? & Standardized classes? & Sim and Real synced? \\
            \midrule
            Oakland 3D \cite{Munoz-2009-10227}   &   2009   &   MLS   &   real   &   44   & \greencheck   &  \redxmark  &  \redxmark  & - \\
            Sydney Urban Objects Dataset \cite{SydneyDatasetde2013unsupervised}  &   2013   &   MLS   &   real   &   26   & \greencheck &  \redxmark  & \redxmark & - \\
            Paris-rue-Madame database \cite{serna2014parisMadame}   &   2014   &   MLS   &   real   &   27   & \greencheck   &   \redxmark  & \redxmark & -   \\
            iQumulus  \cite{vallet2015terramobilita} &   2015   &   MLS   &   real   &   8   & \greencheck &   \redxmark  & \redxmark &  - \\
            TUM-MLS-2016 \cite{zhu_tum-mls-2016_2020}  &   2016   &   MLS   &   real   &   9   & \greencheck &   \redxmark  & \redxmark &  -  \\
            semantic3D.net \cite{hackel2017semantic3d}  &   2017   &   TLS   &   real   &   9   & \greencheck &   \redxmark  & \redxmark &  -  \\
            Paris-Lille-3D \cite{roynard2018parisLille}  &   2018   &   MLS   &   real   &   50   & \greencheck &  \redxmark  & \redxmark &  -  \\
            SynthCity \cite{griffiths2019synthcity}   &   2019   &   MLS   &   simulated   &   9   & \redxmark &   \redxmark  & \redxmark &  -  \\
            A2D2 \cite{geyer2020a2d2}  &   2020   &   MLS   &   real   &   38   & \greencheck &   \redxmark  & \redxmark &  -  \\
            ArCH \cite{matrone2020comparing}  &   2020   &   TLS/MLS/UAV   &   real   &   10   & \greencheck &   \redxmark & \greencheck &  -   \\
            Toronto-3D \cite{tan2020toronto}  &   2020   &   MLS   &   real   &   8   & \greencheck &   \redxmark & \redxmark &  -   \\
            KITTI-360 \cite{liao2021kitti}  &   2021   &   MLS   &   real   &   19   & \greencheck  &   \redxmark  & \redxmark  & - \\
            Paris-CARLA-3D \cite{carla_paris}  &   2021   &   MLS   &   real \raisebox{0ex}{+} simulated   &   23   & $\thicksim$ &  $\thicksim$  & \redxmark & \redxmark \\
            TUM-FAÇADE \cite{tumfacadePaper}  &   2022   &   MLS   &   real   &   17   & \greencheck &  \redxmark & \greencheck & -  \\
            HelixNet \cite{helixnet}  &   2022   &   MLS   &   real   &   9   & \greencheck  &  \redxmark  & \redxmark & - \\
            SUD \cite{SUDdata_SiliviaGonzalez}  &   2023   &   MLS   &   real   &   8   & \greencheck &   \redxmark  & \redxmark &  -  \\
            DELIVER \cite{liao2025benchmarking} &   2025   &   MLS   &   simulated   &  25  & \redxmark &   \greencheck  & \redxmark & - \\
            ZAHA \cite{wysocki2025zaha} &   2025   &   MLS   &   real   &   15   & \greencheck &   \redxmark  & \greencheck & - \\ \hline
            
            \textbf{TrueCity (ours)} &   2025   &   MLS   &   real \raisebox{0ex}{+} simulated   &   12   & \greencheck &   \greencheck  & \greencheck & \greencheck \\
            \bottomrule
        \end{tabular}
    \end{adjustbox}
    \caption{Point cloud benchmark datasets for urban semantic segmentation.} 
    \label{tab:bigTable}
\end{table*}
Another challenge lies in the high variation of class definitions across urban segmentation datasets, which hinders unified, consistent, and fair comparisons of algorithmic performance:
The minimum number is eight and the maximum is 50 in the analyzed datasets (\cref{tab:bigTable}).
The lack of standardized class representations also poses difficulties for developing robust transfer learning approaches.
In related fields such as geomatics, architecture, and civil engineering, international standardization bodies have long established urban object taxonomies for modeling \cite{citygml2objv2,asamOpenDRIVEV1User2015,wysocki2025zaha}. 
These standardized class definitions, however, remain largely overlooked in the design of point cloud segmentation datasets, which limits the direct applicability of their outputs to standardized semantic modeling workflows.
Here, the notable exceptions are ZAHA \cite{wysocki2025zaha}, TUM-FAÇADE \cite{tumfacadePaper}, and ArCH \cite{archDatasetPaper}, grounding their classes in international standards.
Yet, they focus on either facade-only classes (ZAHA, TUM-FAÇADE), excluding any other urban classes; or on cultural-heritage-specific classes (ArCH), limits their application to generic urban scenarios. 
Worth noting are also works focusing on generating synthetic images from virtual 3D assets, yet they are out of scope for the 3D point cloud benchmarks \cite{gaidon2016virtual,nikolenko2021synthetic}.

\subsection{Standardized Semantic City Models}
\label{sec:standards}

For the representation of cities and landscapes in terms of their semantics, geometry, topology, and appearance, the international standard CityGML has become widely adopted by municipalities and entire countries \cite{grogerOGCCityGeography2012,wysockiReviewingOpenData2024}.
CityGML version 2.0 was released by the \gls{OGC} in 2012 and defines a conceptual data model, which is based on the standards from the ISO 191XX series of geographic information standards and therefore supported by geographic information systems.
The standard specifies concepts and class definitions for representing buildings, bridges, tunnels, vegetation, city furniture, and transportation infrastructure across different \glspl{LoD} \cite{grogerOGCCityGeography2012}.
Combined with buildings, detailed facades, vegetation, and city furniture, the street space can be represented in a comprehensive, semantic, and interoperable manner.
As of 2024, there are approximately 216.5 million building models available worldwide as open CityGML datasets \citep{wysockiReviewingOpenData2024}.
This includes the building stocks in \gls{LoD}\,2 of Germany, Switzerland, Poland, and large parts of Japan, which are provided and maintained by public authorities with stable object identifiers.
Moreover, road networks and roadside objects can also be modeled in the OpenDRIVE standard for high-definition mapping, traffic and driving simulation applications \cite{asamOpenDRIVEV1User2015,schwabRequirementAnalysis3D2019}.
Objects defined in OpenDRIVE using parametric geometries can be transformed into CityGML’s explicit geometries through a dedicated conversion method \cite{schwab2020spatio}.


\section{TrueCity Benchmark Dataset}
\label{sec:ingol}

\subsection{3D Semantic Road Space Classes}

As shown in Table~\ref{tab:bigTable}, there is a scarcity of point cloud benchmark datasets capturing the same environment through both real and simulated laser scanning.
Addressing this scarcity requires an environment model with semantic information to enable the derivation of semantically labeled point clouds from laser scanning simulation.
To ensure compatibility with established standardized data models, we propose a class list of 12 classes harmonized with the standards CityGML 2.0 \cite{grogerOGCCityGeography2012} and OpenDRIVE 1.4 \cite{asamOpenDRIVEV1User2015}.
Leveraging standardized class definitions facilitates seamless integration and reuse in downstream methods and applications, such as 3D road space and facade semantic reconstruction \cite{González-Collazo21032025,wysocki2023scan2lod3}.
A detailed description of these classes is provided in Table~\ref{tab:class-list}, along with their corresponding classes defined in the standards.
The standards further specify type categories, functions, and usage lists for objects; for details see \cite{grogerOGCCityGeography2012,asamOpenDRIVEV1User2015}.
\begin{table*}[htb]
    \newcommand{\hlineafter}{%
    }
    \newcommand{\colorBox}[1]{%
        {\color{#1}\rule[-0.3ex]{1em}{1em}}%
    }
    \setlength{\tabcolsep}{2pt}
    \footnotesize
    \centering
    \begin{tabular}{rlllll}
    \toprule
    \multirow{2}{*}{\#} & & \multirow{2}{*}{Class} & \multirow{2}{*}{Description} & \multicolumn{2}{c}{Corresponding standard class} \\
    \cmidrule(lr){5-6}
    & & & & CityGML 2.0 \citep{grogerOGCCityGeography2012} & OpenDRIVE 1.4 \cite{asamOpenDRIVEV1User2015} \\
    \midrule
     1 & \colorBox{semanticLabelRoadSurface} & RoadSurface & Vehicle-allowed surfaces without sidewalks & (Auxiliary)TrafficArea & LaneSectionLRLane \\
     \hlineafter
     2 & \colorBox{semanticLabelGroundSurface} & GroundSurface & Pedestrian-allowed surfaces without road surface & (Auxiliary)TrafficArea, OuterFloorSurface & LaneSectionLRLane \\
     \hlineafter
     3 &  \colorBox{semanticLabelCityFurniture} & CityFurniture & Vertical urban installation without building-attached objects & CityFurniture & Signal, RoadObject \\
     \hlineafter
     4 & \colorBox{semanticLabelVehicle} & Vehicle & Parked or moving vehicles & – & – \\
     \hlineafter
     5 & \colorBox{semanticLabelPedestrian} & Pedestrian & Standing or moving persons & – & – \\
     \hlineafter
     6 & \colorBox{semanticLabelWallSurface} & WallSurface & Building parts without roofs, installations, facade elements & WallSurface & RoadObject \\
     \hlineafter
     7 & \colorBox{semanticLabelRoofSurface} & RoofSurface & Building parts forming roof structures & RoofSurface & RoadObject \\
     \hlineafter
     8 & \colorBox{semanticLabelDoor} & Door & Openings allowing entering objects with gates & Door & – \\
     \hlineafter
     9 & \colorBox{semanticLabelWindow} & Window & Openings and its outer blinds without entries & Window & – \\
     \hlineafter
     10 & \colorBox{semanticLabelBuildingInstallation} & BuildingInstallation & Building-attached installation & BuildingInstallation, OuterCeilingSurface & – \\
     \hlineafter
     11 & \colorBox{semanticLabelSolitaryVegetationObject} & SolitaryVegetationObject & Vegetation with tree trunks and branches & SolitaryVegetationObject & RoadObject  \\
     \hlineafter
     12 & \colorBox{semanticLabelNoise} & Noise & Noisy points and any other non-annotated element & – & – \\
     \bottomrule
    \end{tabular}
    \caption{Semantic road space classes harmonized with the class definitions of the standards CityGML 2.0 and OpenDRIVE 1.4.}
    \label{tab:class-list}
\end{table*}

\begin{figure*}[htb]
    \centering
    \includegraphics[width=\linewidth]{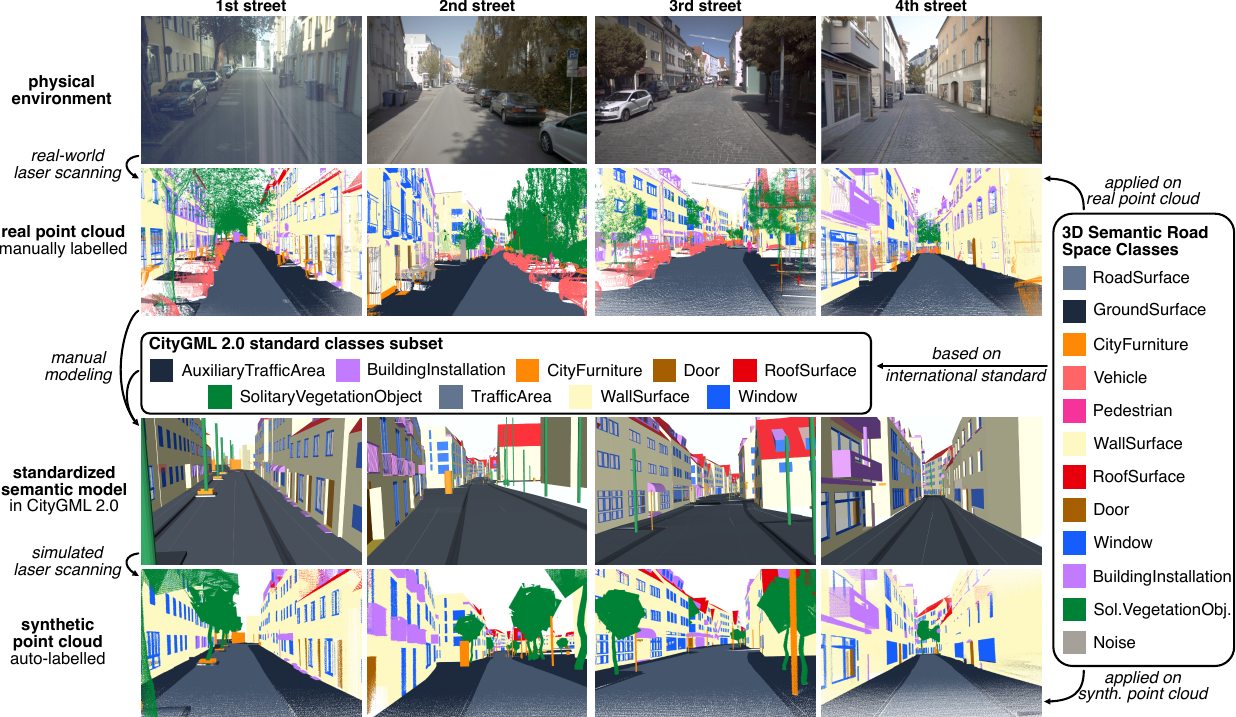}
    \caption{Real-world point cloud (2nd row), which was manually labeled according to the class list of Table \ref{tab:class-list}, used for manual modeling of semantic 3D models (3rd row), which in turn were used to simulate and auto-label synthetic point clouds (4th row).} 
    \label{fig:benchmarkOverview}
\end{figure*}
\subsection{Real-World Data Acquisition}
For the physical environment of the TrueCity benchmark dataset, we selected the inner city of Ingolstadt, a mid-sized German town with a dense and challenging urban setting.
Figure~\ref{fig:benchmarkOverview} shows the dataset covers four streets totaling about 500\,m, featuring adjacent facades of 2-4-story buildings, vegetation, street furniture, vehicles, and pedestrians.

\noindent\textbf{Real-World Laser Scanning}
The high-precision laser scans are conducted by the company 3D Mapping Solutions using their Mobile-Road-Mapping-System (MoSES), which is mounted on a minivan to collect the real-world point clouds \cite{3D_mapping,haigermoserRoadTrackIrregularities2015}. 
In total, 113 million points are recorded, with densities of up to 3,000\,pts/m\textsuperscript{2}, while the setup achieves a relative accuracy of 1-3\,cm.
The inertial measurement unit is complemented by odometer sensors and a high-precision differential GPS with \gls{RTK} correction data from the German satellite positioning service (SAPOS), ensuring accurate georeferencing \cite{haigermoserRoadTrackIrregularities2015}.
The georeferenced real-world point clouds are provided in the projected \gls{CRS} UTM Zone 32N (EPSG:25832).

\begin{figure}[htb]
    \centering
    \includegraphics[width=\linewidth]{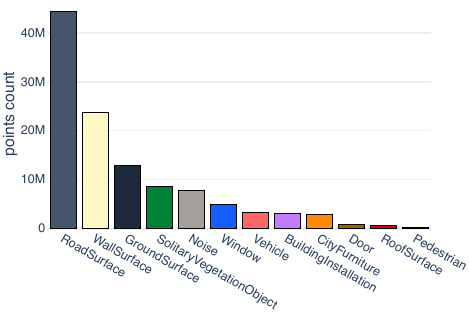}
    \caption{TrueCity represents the typical long-tail distribution challenge of real-world data.} 
    \label{fig:realPointDistribution}
\end{figure}

\noindent\textbf{Semantic Annotation}
Our labeling of real-world point clouds follows a three-step process.
First, we divide the point cloud into parts of connected components based on their spatial distance.
Second, the ground surfaces, including roads and sidewalks, are separated from elevated objects using the cloth simulation filtering (CSF) algorithm \cite{CSF}. 
Third, we manually select the corresponding subset of points and assign it a class ID from Table~\ref{tab:class-list}.
Finally, all point clouds are merged, resulting in the points-per-class distribution shown in Figure~\ref{fig:realPointDistribution}. 
This phenomenon posits TrueCity in the realistic long-tail distribution regime, typical challenge of real-world data \cite{wysocki2025zaha,dai2017scannet}.

\subsection{Synthetic Data Acquisition}
\noindent\textbf{Semantic 3D City Model}
To achieve precise synchronization, we manually model the buildings and detailed facades using the acquired real point clouds, in accordance with the CityGML standard.
The road network and roadside objects are manually curated using the same point clouds 
as an OpenDRIVE dataset, which we subsequently convert to CityGML 2.0 \cite{schwab2020spatio,schwabValidationParametricOpenDRIVE2022}.
This yields a comprehensive, georeferenced, and semantic model of the four streets, shown in the third row of Figure~\ref{fig:benchmarkOverview}.
We procedurally replace the coarse geometric representation of trees with detailed 3D assets and translate all 3D models into
a local \gls{CRS} for the laser scanning simulation, while retaining the class information.

\noindent\textbf{Simulated Laser Scanning}
To replicate the real-world laser scanning process, we use the CARLA driving simulator, which includes a configurable LiDAR sensor model \cite{dosovitskiy2017carla}.
We resimulate the real-world sensor data collection route using the LiDAR parameters specified in Table~\ref{tab:lidar-configuration}.
To further enhance realism, we incorporate a Gaussian-distributed range error proposed by \cite{spiegel2021} as follows:
\begin{equation}
    r' = r + \varepsilon, \quad 
    \varepsilon \sim \mathcal{N}(0, \rho^2), \quad 
    \rho = 0.02\,\text{m}
\end{equation}
The simulation accounts only for static environment objects; consequently, the synthetic point clouds exclude the \emph{Vehicle} and \emph{Pedestrian} classes.

\section{Experiments}
\label{sec:experiments}
We evaluate on TrueCity under controlled synthetic-real (\%S--\%R) mixtures of 100S--0R, 75S--25R, 50S--50R, 25S--75R, and 0S--100R. We form each S--R mixture by assigning contiguous spatial segments to the synthetic or real domain, rather than interleaving points as explored in our parallel submission~\cite{ourPrevious}.
The test and validation data remain real-only throughout the experiments.
For a target real fraction $r$, we allocate $r$ of the street-surface area and match the total street length across mixtures to isolate composition effects. Street length and spatial coverage are fixed across mixtures, only the raw point totals vary modestly (112.9-137.6M) due to domain-specific sampling densities. The detailed class distribution in each mixture can be found in Supplement Sec.~\ref{sec:experiment_detail_result}.

\begin{figure}[h]
    \centering
    \includegraphics[width=\columnwidth]{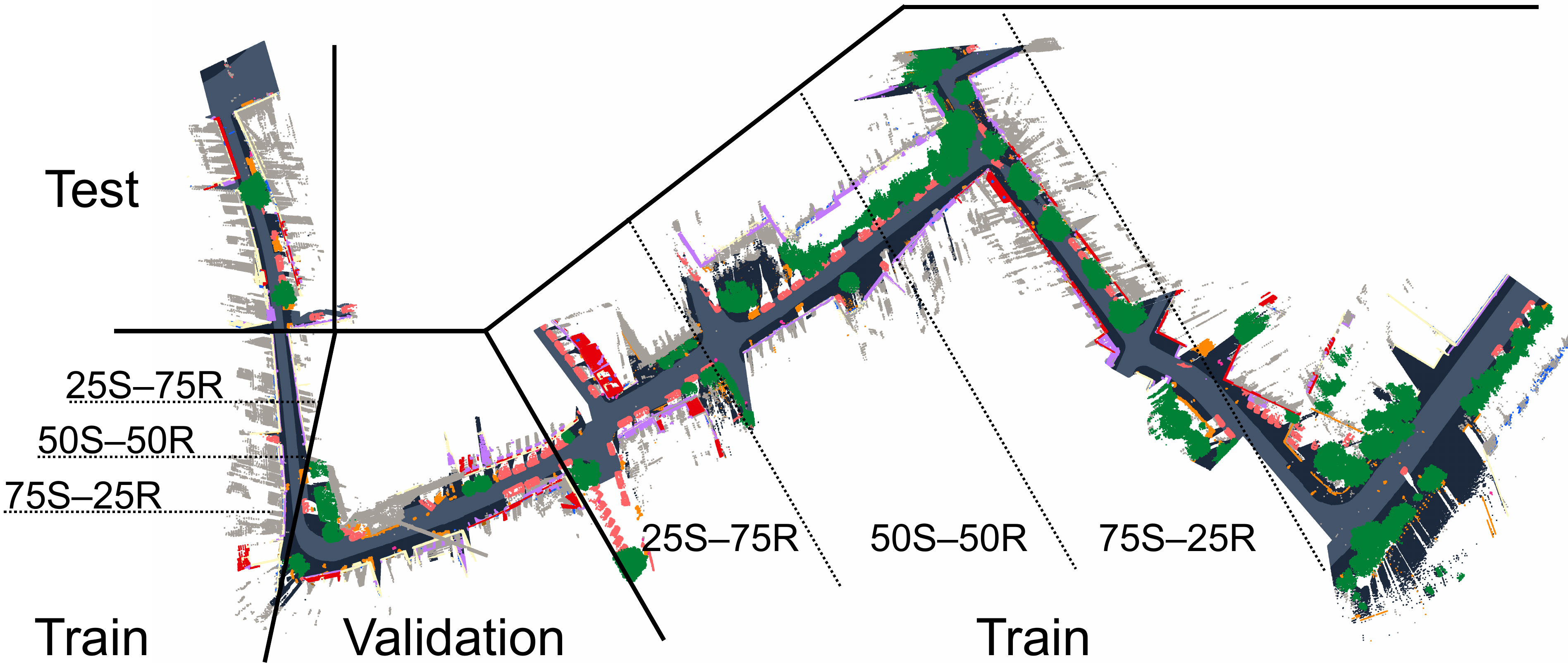}
        \caption{Top-down schematic of S--R mixtures along a continuous streetscape. Solid lines mark train/validation/test splits; dashed lines mark boundaries between contiguous synthetic and real segments for each mixture ratio.}
    \label{fig:data_split}
\end{figure}

\noindent \textbf{Evaluation protocol}
All results are evaluated on a fixed real-only validation/test split. Synthetic scans appear only in training when included by the mix. We report mIoU, OA, and mAcc over the 12-class TrueCity taxonomy. Aggregate results appear in Table~\ref{tab:iou_partitions}. A per-class IoU breakdown for PointNet++, KPConv, and Point Transformer v1 appears in Table~\ref{tab:mIoU_per_class}, with full results in the Supplement (Sec.~\ref {sec:experiment_detail_result}).

\subsection{Baseline Semantic Segmentation Methods}
To probe synthetic-real (S--R) domain shift on TrueCity, we evaluate a representative suite of point-cloud semantic segmentation baselines widely used for urban scenes and consistently strong on established benchmarks \cite{semanticKITTI,nuscenes,tan2020toronto,carla_paris}. The suite spans three complementary families: point-based, kernel-based, and transformer-based, so that trends are not tied to a single inductive prior. 
See the Supplement for the detailed description (Sec.~\ref {sec:experiment_detail_result})

\section{Results and Discussion}
\label{sec:discussion}
\begin{table*}[t]
  \centering
  \small
  \setlength{\tabcolsep}{3.5pt}
  \begin{tabularx}{\textwidth}{@{}l*{10}{>{\centering\arraybackslash}X}@{}}
    \toprule
    \multirow{2}{*}{Model} &
    \multicolumn{2}{c}{100S--0R} &
    \multicolumn{2}{c}{75S--25R} &
    \multicolumn{2}{c}{50S--50R} &
    \multicolumn{2}{c}{25S--75R} &
    \multicolumn{2}{c}{0S--100R} \\
    \cmidrule(lr){2-3}\cmidrule(lr){4-5}\cmidrule(lr){6-7}\cmidrule(lr){8-9}\cmidrule(lr){10-11}
    & mIoU & OA & mIoU & OA & mIoU & OA & mIoU & OA & mIoU & OA \\
    \midrule
    PointNet~\cite{qi2017pointnet}                & 6.03 & 30.36 & 10.74 & 48.10 & 10.89 & 49.29 & 13.10 & 47.99 & \textbf{14.51} & \textbf{49.82} \\
    PointNet++~\cite{qi2017pointnet++}              & 9.72 & 34.39 & 20.95 & 62.80 & 23.18 & \textbf{65.36} & \textbf{25.38} & 63.27 & 23.39 & 63.15 \\
    RandLA-Net~\cite{hu2020randlanet}              &  8.98 & 35.40 & 13.25 & 50.32 & 15.73 & \textbf{59.37} & 16.89 & 57.09 & \textbf{17.71} & 54.57 \\
    \midrule
    KPConv~\cite{thomas2019kpconv}                  & 15.84 & 50.07 & 21.55 & 62.08 & 28.50 & 61.62 & 22.33 & 61.92 & \textbf{29.90} & \textbf{62.80}  \\
    \midrule
    Point Transformer v1~\cite{zhao2021point}    & 16.30 & 57.54 & 19.79 & 60.29 & 23.43 & 67.54 & 24.66 & \textbf{68.70} & \textbf{28.89} & 67.98 \\
    Point Transformer v3~\cite{Wu_2024_pointtrasformerv3}    & 14.13 & 53.15 & 19.29 & 60.22 & \textbf{25.30} & \textbf{65.94} & 24.64 & 65.72 & 25.24 & 60.75 \\
    Superpoint Transformer~\cite{robert2023spt}  & 14.31 & 54.17 & 17.01 & \textbf{58.62} & 14.22 & 54.63 & 19.61 & \textbf{56.98} & 15.96 & 54.64 \\
    OctFormer~\cite{wang2023octformer}               & 13.07 & 53.30 & 14.17 & 55.34 & 14.22 & 49.71 & 13.91 & 50.97 & \textbf{17.65} & \textbf{56.28} \\
    \bottomrule
  \end{tabularx}
    \caption{TrueCity segmentation under synthetic-real (S--R) mixes. We report mIoU and OA; shifts along S--R reveal family-specific inductive biases in point-, kernel-, and transformer-based models. Bold marks the best value for each model across mixtures.}

  \label{tab:iou_partitions}
\end{table*}

\begin{figure*}[htb]
    \centering
    \includegraphics[width=\textwidth]{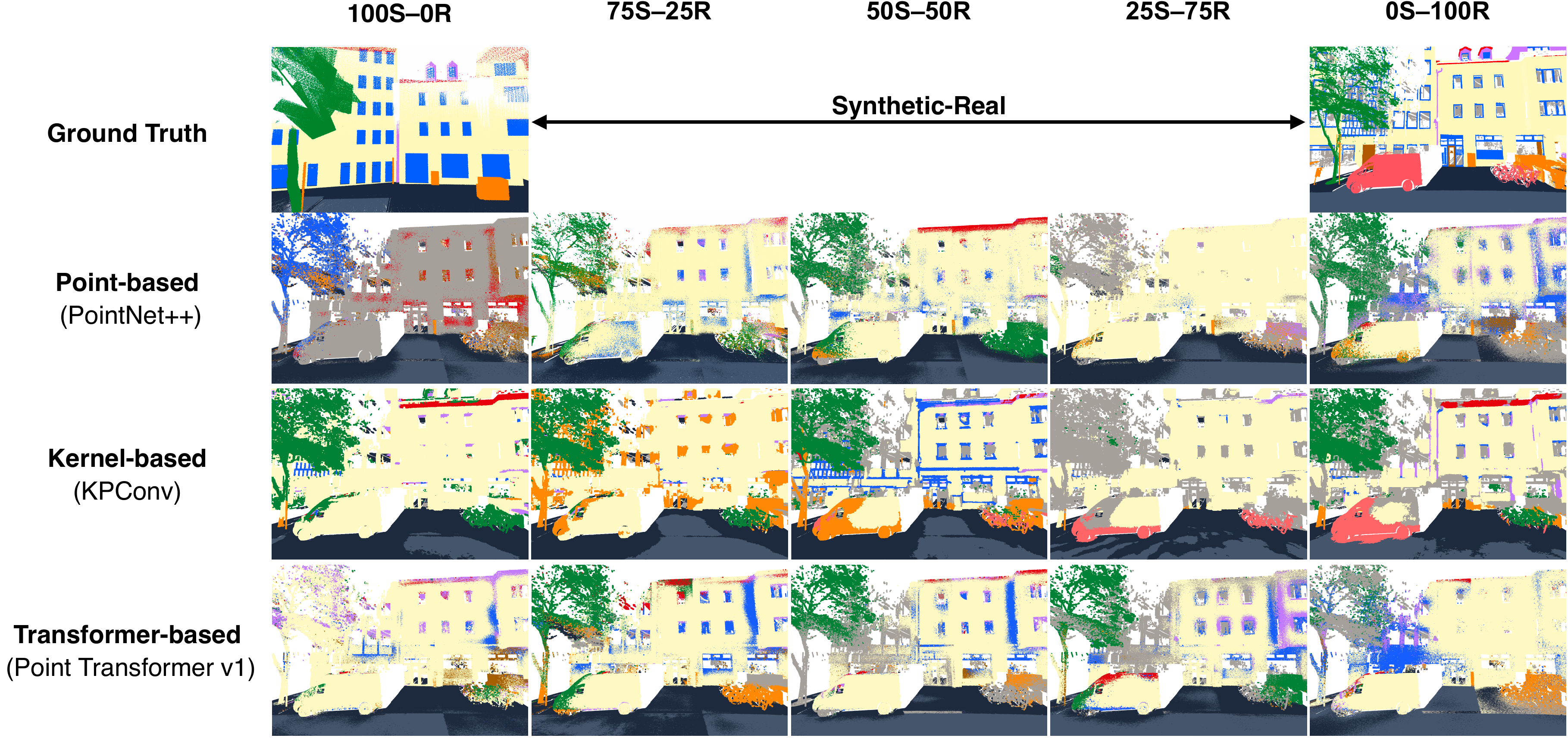}
\caption{Qualitative impact of the synthetic–real (S--R) training mix on models from different methods (Point-based, Kernel-based and Transformer-based). We also present the ground truth synthetic and real point clouds; colors follow the TrueCity legend.}
\label{fig:experiment_visualize}
\end{figure*}

Synthetic data helps, but its utility depends on the model’s inductive bias. Architectures with global attention handle synthetic well: Point Transformer v3 improves mIoU from 14.13\% at 100S--0R to broad maximum of 25.30\% at 50S--50R and 25.24\% at 0S--100R. Point Transformer v1 benefits primarily from real data, moving from 16.30\% at 100S--0R to 28.89\% at 0S--100R, indicating a stronger reliance on sensor statistics. Hierarchical set abstraction also profits from a modest synthetic prior: PointNet++ peaks at 25.38\% with 25S--75R versus 23.39\% with 0S--100R, likely because synthetic coverage reduces sparsity while real data calibrates noise. Methods driven by local neighborhoods or superpoints lean heavily on real data: RandLA-Net rises from 8.98\% to 17.71\%, and Superpoint Transformer from 14.31\% to 19.61\% as the real fraction increases, consistent with their dependence on LiDAR sampling artifacts underrepresented in simulation. The practical recipe is straightforward: use synthetic as a patch, not a replacement, favor balanced mixes (around 50S--50R or 25S--75R) for strong transformers, and bias toward real data for locality-driven models while using synchronized synthetic twins to fill occlusions and thin structures.

\subsection{Insignificant Domain Gap Classes} 

Table~\ref{tab:mIoU_per_class} highlights that some classes show minimal sensitivity to the proportion of real data, either maintaining strong IoU across all settings or reaching near-peak performance with limited real supervision. 

\textit{WallSurface} is well captured synthetically by all three models.  
Even without real data, Point Transformer v1 reaches 53.15\% IoU and improves only modestly to 67.10\% with full real supervision, a relative gain of 26.2\%.  
KPConv shows a similar trend, attaining 73.55\% with 75\% real data, which corresponds to a 22.6\% improvement over the synthetic-only setting (59.97\%).  
In contrast, PointNet++ relies heavily on local sampling and interpolation. With only synthetic input, it performs poorly (0.20\%), but a small fraction of real data (25\%) boosts its performance dramatically to 56.90\%.  
After this sharp correction, further gains remain limited.  
This confirms that large planar facades with consistent geometry can already be recognized effectively using the synthetic data.  
For Point Transformer v1 and KPConv, real data provides only incremental refinement, whereas PointNet++ requires minimal real supervision to overcome its synthetic-domain weakness.  

\textit{RoadSurface} also starts from a strong synthetic baseline: 60.90\% for PointNet++ and 62.41\% for Point Transformer v1.  
Both reach around 80\% with 50\% real data, corresponding to relative gains of 33.3\% and 28.7\%.  
In contrast, KPConv starts much lower at 30.64\% but rises steeply to 69.26\% with just 25\% real data, a 126.1\% improvement. 
This indicates that convolutional kernels rely more heavily on realistic geometrical cues, whereas PointNet++ and Point Transformer v1 are already robust under synthetic supervision.  
\textit{GroundSurface} records lower absolute scores overall, but all three models show a steady upward trajectory, which again reflects moderate domain sensitivity.  

\textit{SolitaryVegetationObject} shows consistent trends across methods: Point Transformer v1 improves from 15.54\% to 35.35\% with 25\% real data, KPConv rises from 45.74\% to 70.51\%, and PointNet++ from 0\% to 50.60\%. 
Although performance continues to increase with more real supervision, the early gains highlight that a small fraction of real data is sufficient to close much of the initial domain gap. 

Overall, these categories share traits of geometric regularity and relatively simple structural context. 
Synthetic data provides a strong baseline, while real data mainly refines fine appearance cues. 
Even small proportions of real data suffice to approach optimal performance, making these classes representative of insignificant domain gap behavior.

\begin{table*}[htb]
  \centering
  \small
  \setlength{\tabcolsep}{4pt}
  \begin{tabular}{lccccccccccccccc}
    \toprule
    \multirow{2}{*}{Class} &
      \multicolumn{5}{c}{PointNet++} &
      \multicolumn{5}{c}{KPConv} &
      \multicolumn{5}{c}{Point Transformer v1} \\
    \cmidrule(lr){2-6}\cmidrule(lr){7-11}\cmidrule(lr){12-16}
    & 100S & 75S & 50S & 25S & 0S
    & 100S & 75S & 50S & 25S & 0S
    & 100S & 75S & 50S & 25S & 0S \\

    \midrule
    RoadSurface              & 60.90 & 73.90 & \textbf{81.20} & 80.30 & 72.50 & 30.64  & \textbf{69.26}  & 54.26  & 56.31  & 54.59 & 62.41 & 61.89 & \textbf{80.33} & 77.93 & 70.44 \\
    GroundSurface            & 33.20 & 45.30 & \textbf{50.40} & 49.60 & 32.40 & 24.67  & \textbf{36.49}  & 31.91  & 33.20  & 35.96 & 31.94 & 37.54 & \textbf{45.54} & 37.09 & 41.67 \\
    CityFurniture            & 15.20 & 14.60 & 21.30 & 25.50 & \textbf{46.20} & \textbf{18.25}  & 10.09  & 12.67  & 1.85  & 14.57 & 25.89 & 19.50 & 13.51 & 14.23 & \textbf{44.83} \\
    Vehicle            & \textbf{3.50}   &   0.00   &   0.18   &   0.51   &   0.01 &  0.00 &  0.00 & 38.93 & \textbf{70.33} & 63.64 & 0.00 & 0.51 & 4.41 & 1.51 & \textbf{6.54} \\
    Pedestrian            &  0.00   &   0.00   &   0.00   &  \textbf{0.32}   &   0.04  & 0.00 & 0.00 & 0.00 & 0.00 & 0.00 & 0.00 & 0.00 & \textbf{0.05} & 0.00 & 0.00 \\
    WallSurface              & 0.20 & 56.90 & 63.50 & \textbf{65.70} & 63.20 & 59.97  & 63.21  & 68.54  & \textbf{73.55}  & 71.03 & 53.15 & 61.42 & 64.39 & 65.32 & \textbf{67.10} \\
    RoofSurface              & 0.00  & \textbf{0.80}  & 0.30  & 0.10  & 0.30  & 0.82  & 0.01  & 0.47  & 0.00  & \textbf{1.79}  & 0.10  & 0.29  & \textbf{0.92}  & 0.30  & 0.08  \\
    Door                     & 0.00  & 0.00  & 0.20  & 0.10  &\textbf{0.70}  & 0.00  & 0.00  & 0.00  & 0.00  & 0.00 & 0.01  & 0.13  & 0.51  & \textbf{0.66}  & 0.01  \\
    Window                   & 3.61  & 7.30  & 4.50  & 4.20  & \textbf{14.60} & 1.86  & 0.43  & \textbf{18.21}  & 1.17  & 2.89  & 3.28  & 5.61  & 7.61  & 11.78 & \textbf{31.29} \\
    BuildingInstallation & 0.00  & 0.70  & 3.00  & 1.70  & \textbf{4.60}  & \textbf{7.85}  & 3.02  & 0.85  & 0.85  & 3.72  & 3.08  & 5.24  & 6.69  & \textbf{8.87}  & 3.41  \\
    Sol.VegetationObj. & 0.00 & \textbf{50.60} & 39.30 & 45.60 & 14.40 & 45.74  & 70.51  & \textbf{81.45}  & 0.41  & 77.13 & 15.54 & 35.35 & 34.70 & 43.22 & \textbf{52.31} \\
    Noise                    & 0.00  & 1.30  & 14.30 & 30.90 & \textbf{31.70} & 0.31  & 5.58  & \textbf{34.73}  & 30.26  & 33.43 & 0.16  & 9.98  & 22.43 & \textbf{35.04} & 28.96 \\
    \bottomrule
  \end{tabular}
    \caption{Per-class IoU (↑) for one representative per family: PointNet++ (Point-based), KPConv (Kernel-based), and Point Transformer v1 (Transformer-based) across synthetic-real (S--R) training mixes. Columns 100S, 75S, 50S, 25S, and 0S denote the synthetic fraction used in training (e.g., 100S--0R). Evaluation is on the fixed test split. Bold marks the best value for each model across mixtures. Full per-class tables appear in the Supplementary Material (Sec.~\ref{sec:experiment_detail_result}).}

  \label{tab:mIoU_per_class}
\end{table*}

\subsection{Significant Domain Gap Classes}

As shown in Table~\ref{tab:mIoU_per_class}, several categories display substantial domain gaps, where performance depends heavily on the balance of synthetic and real supervision.

\textit{Door}, \textit{RoofSurface}, and \textit{BuildingInstallation} remain the weakest classes across all methods, with IoU rarely exceeding 8\%. 
Adding real data can increase scores by several orders of magnitude (e.g., Point Transformer v1 rises from 0.01\% to 0.66\% for \textit{Door}), yet these gains are unstable and often collapse when real data dominates. 
These categories depend on fine-scale geometry and contextual cues absent from simulation, while limited real samples risk overfitting to narrow geometrical patterns.

\textit{Noise} shows a more consistent improvement: PointNet++ climbs from 0\% to 31.70\%, KPConv from 0.31\% to 34.73\%, and Point Transformer v1 from 0.16\% to 35.04\%, corresponding to an increase of over 30\% in the best case.  
Although Point Transformer v1 drops slightly when trained without synthetic data, all three methods display a strong positive trend, confirming that real-world variation is essential for learning this diverse class.

\textit{CityFurniture} overall exhibits a U-shaped trend across all models. 
Performance drops sharply at intermediate real-data proportions (e.g., KPConv declines from 18.25\% to 1.85\%), but recovers once training relies entirely on real supervision, reaching 14.57\%. 
This suggests that mid-range ratios trigger domain conflicts, where synthetic and real samples provide competing cues, whereas fully real data resolves this inconsistency.

\textit{Window} improves steadily for PointNet++ (3.61\% $\rightarrow$ 14.60\%) and Point Transformer v1 (3.28\% $\rightarrow$ 31.29\%), yet absolute scores remain modest. 
Their small size, thin geometry, and similarity to walls, combined with reflections and occlusions, make them difficult to segment. 
Synthetic data fails to capture these effects, and limited real samples cannot cover the variability, constraining performance.

Overall, for large domain gap classes, which often involve fine-scale geometry, occlusions, or high variability in appearance, finding the right synthetic-real balance is essential for stable and accurate segmentation.

\subsection{Limitations and Future Work}

TrueCity provides a unique combination of synchronized a) real-world cm-accurate captured point clouds, b) semantic 3D city models, and c) annotated and simulated point clouds in the same real-world location.
Our comprehensive experiments were conducted on geometric values, without mixing them with any radiometric values to ensure fair comparison across three data subsets on the geometry level.
Yet, the radiometry can also be included by, for example, image projection as we attach images and their trajectories taken during the mobile mapping acquisition.
Owing to the high acquisition cost and data accuracy, we acknowledge that such a dataset cannot reach the scalability rate of image-based datasets, as noted in other well-established point cloud datasets, e.g., ScanNet \cite{dai2017scannet}.
Yet, the unprecedented accuracy and quality should present the significant merit of this dataset.
In TrueCity, the focus lies on static objects, as the real-world capture is collected at one timestamp, and introducing de-synchronized dynamic objects would impede per-class domain shift analysis.
Nevertheless, since we provide all three subsets of data, dynamic objects can be simulated in various traffic scenarios.
Further elaboration is presented in in the Supplement.

\section{Conclusion}
\label{sec:conclusion}

In this paper, we present TrueCity, the benchmark dataset comprising real and simulated point clouds synchronized with underlying semantic 3D city models representing the same geographic location.

Based on our extensive experiments, we conclude that TrueCity enables in-depth coherent analysis of the domain gap across different network architectures, removing the uncertainty factor stemming from fictitious, de-synchronized scenes.  
Another finding of this study indicates that real point clouds can be partially replaced by simulated ones without compromising performance: several methods achieve comparable or even superior results with 50\% real data compared to full real training. 
For example, Point Transformer v3 achieves 65.94\% OA with 50\% synthetic data, an 8.5\% relative improvement over the all-real setting (60.75\%).

Also, we observe that material complex classes suffer from simplistic assumptions in simulators (e.g., rays not penetrating through glass). For instance, without real data, \textit{Window} IoU is only 3.28\% for Point Transformer v1 and 3.61\% for PointNet++, showing the limits of synthetic data. 
In contrast, well-represented and primitive-like classes can be simulated to a large extent (e.g. \emph{WallSurface} reaches 59.97\% IoU for KPConv trained on synthetic data only). 
Finally, complex non-manmade classes such as \textit{SolitaryVegetationObject} remain difficult to synthesize owing to oversimplified representations in the underlying 3D model. Yet, even small amounts of real data bring substantial improvements, e.g., PointNet++ improves from 0.00\% at 100S--0R to 50.60\% at 75S--25R. 
We are convinced this dataset can foster further research on cross-domain gap analysis and large-scale data simulation.

\section*{Acknowledgements}
\label{sec:acknowledgements}

We would like to thank the company 3D Mapping Solutions for providing the real-world MLS point clouds.
We are also grateful to the Munich Data Science Institute (MDSI) and Dr. Ricardo Acevedo Cabra for his support through the TUM Data Innovation Lab.
Finally, we thank the Data Innovation Lab members — Patrick Madlindl, Xinyuan Zhu, and Florian Hauck — for their valuable contributions.

This work is supported by the German Federal Ministry of Transport and Digital Infrastructure (BMVI) within the \textit{Automated and Connected Driving} funding program under the Grant No. 01MM20012K (SAVeNoW).
Duc Nguyen is supported by the DAAD program Konrad Zuse Schools of Excellence in Artificial Intelligence, sponsored by the German Federal Ministry of Education and Research.

\newpage
{
    \small
    \bibliographystyle{ieeenat_fullname}
    \bibliography{main}
}
\clearpage
\setcounter{page}{1}
\maketitlesupplementary
Here, we further discuss our experiments. 
Complementary to this supplementary materials is our submission here: \cite{ourPrevious}, which describes creation of the simulation environment.
\section{Laser Scanning Simulation}
\label{sec:simulation_suppl}
\subsection{Configuration}
For the laser scanning simulation, we configure the sensor to match the real-world laser scanning setup.
\begin{table}[htb]
     \centering
     \begin{tabular}{l r}
         \toprule
         Parameter & Value \\
         \midrule
         Number of lasers & $128$ \\
         Points generated by all lasers & $500{,}000$\ points/s \\
         Rotation frequency & $20$\,Hz \\
         Upper FOV & $15^\circ$ \\
         Lower FOV & $-25^\circ$ \\
         Horizontal FOV & $360^\circ$ \\
         Range & $100$\,m \\
         \bottomrule
     \end{tabular}
     \caption{Configuration parameters of the LiDAR sensor model.}
     \label{tab:lidar-configuration}
\end{table}

\subsection{Radiometry}
As highlighted in the main body of the paper, we do not include any radiometric features in the simulation (e.g., RGB).
We opt for this approach due to the limited field-of-view of the acquisition camera, which does not fully cover the field-of-view of the laser scanners. 
This issue would result in incomplete and non-robust image-to-model projections.
A possible solution for model texturing is presented in \cite{tang2025texture2lod3}; however, it is subject to wide-angle facade observation, and the choice of the appropriate image for texturing is based on a heuristic, which introduces further randomness to the scan-to-model domain gap analysis.
Nevertheless, we are convinced that future work shall further investigate the impact of radiometry and thus object material on the simulation and resulting domain gap.

\subsection{Dynamic Objects}
The dynamic objects are not included in the simulation setup, as discussed in the main body of the paper.
The large discrepancy between the accurate as-it-happened simulation and actual measurements dictates this choice.
Even if the procedural vehicle and pedestrian models can be placed in a scene, the simulation of their realistic trajectories is questionable and would include large heuristics in the simulation, which would make the domain gap analysis hardly tractable.
Additionally, due to the speed of the mapping vehicle and the observed vehicles and pedestrians, the dynamic object representation is sparse and mainly noisy. We observe the impact of this phenomenon in Tab.~\ref {tab:mIoU_per_class}, where the segmentation results on real-only data also show a large discrepancy between the baseline models.

\section{Experimental Setup}
\label{sec:experiment_setup}

\subsection{Baseline Segmentation Models}

\paragraph{Point-based models}
operate directly on unordered points with permutation-invariant set functions and lightweight neighborhood aggregation. In our experiments, we use PointNet \cite{qi2017pointnet}, PointNet++ \cite{qi2017pointnet++}, and RandLA-Net \cite{hu2020randlanet}. PointNet applies per-point MLPs with a symmetric reduction for global context; PointNet++ adds hierarchical sampling and grouping to capture local structure; RandLA-Net improves scalability via random sampling with attentive local aggregation.
\paragraph{Kernel-based models} 3D convolutions impose locality and yield predictable receptive-field growth. We use KPConv \cite{thomas2019kpconv}, which places learnable kernel points within each neighborhood and aggregates features with geometry-aligned weights in an encoder-decoder pyramid, preserving small-scale structure while expanding context.
\paragraph{Transformer-based models} attention replaces fixed kernels with content-adaptive routing across local neighborhoods and long-range, semantically related regions. We consider region-level attention with Superpoint Transformer \cite{robert2023spt} and octree-structured attention with OctFormer \cite{wang2023octformer}, as well as point-level attention with Point Transformer v1 and v3 \cite{zhao2021point,Wu_2024_pointtrasformerv3}. Superpoint Transformer attends over a superpoint graph \cite{landrieu2018spg}; OctFormer organizes tokens on an octree for multi-scale attention; Point Transformer uses localized self-attention with relative positional encodings, and v3 tightens geometric invariances and reduces overhead for large-scale outdoor LiDAR.

\subsection{Training Setup}
Unless otherwise noted, we maintain a unified training setup: all models are trained for \(\,100\,\) epochs with a constant learning rate of \(10^{-4}\), a mini-batch size of \(32\), and the AdamW optimizer. 
Model-specific deviations from this setup are described in Sec.~\ref{sec:training_recipe}.
Experiments run on NVIDIA H40, L40, and RTX 6000 Ada Generation GPUs; to isolate the effect of the real/synthetic composition, we keep the training dynamics (optimizer, schedule, batch size, total epochs, augmentations, and point budget) identical across conditions and architectures. 
Whenever possible, we rely on the authors’ public implementations with minimal changes. 

\subsection{Evaluation Setup}
Evaluation follows the same point budget and normalization as training. 
We report three standard metrics for semantic segmentation: mean Intersection-over-Union (mIoU), mean Accuracy (mAcc), and Overall Accuracy (OA). 
For $C$ semantic classes, let $n_{ij}$ denote the number of points of class $i$ predicted as class $j$, and $n_{i} = \sum_{j} n_{ij}$ the total number of points in class $i$. 
The metrics are defined as:
\begin{align}
\text{IoU}_{i} &= \frac{n_{ii}}{n_{i} + \sum_{j} n_{ji} - n_{ii}}, \\
\text{mIoU} &= \frac{1}{C} \sum_{i=1}^{C} \text{IoU}_{i}, \\
\text{Acc}_{i} &= \frac{n_{ii}}{n_{i}}, \\
\text{mAcc} &= \frac{1}{C} \sum_{i=1}^{C} \text{Acc}_{i}, \\
\text{OA} &= \frac{\sum_{i=1}^{C} n_{ii}}{\sum_{i=1}^{C} n_{i}}.
\end{align}

\section{Training Recipe}
\label{sec:training_recipe}
\subsection{PointNet and PointNet++}
\noindent Each input point cloud is downsampled to \(2{,}048\) points via farthest-point sampling to preserve geometric coverage. We apply a fixed augmentation regimen: normalization to the unit sphere, a random rotation about the vertical (\(z\)) axis, independent reflections across the \(x\) and \(y\) axes, isotropic scaling drawn from a uniform distribution, and additive jitter. The networks are trained with cross-entropy under the global settings of Sec.~\ref{sec:experiment_setup}.

\subsection{RandLA-Net}
\noindent Training operates on pre-sharded point clouds. For each iteration we form a mini-batch by uniformly sampling a fixed-size subset of \(2{,}048\) points per sample; at evaluation time we sweep scenes with sequential, non-overlapping chunks. Augmentations follow the original recipe of \cite{hu2020randlanet} (normalization, random \(z\)-rotation, axis flips, isotropic scaling, elastic distortion, and jitter). Optimization and schedules are identical to Sec.~\ref{sec:experiment_setup}.

\subsection{KPConv}
\noindent We use the reference KPConv segmentation model \cite{thomas2019kpconv} with the authors’ architectural defaults. To ensure parity across methods, each training item is reduced to \(2{,}048\) points via farthest-point sampling and processed with the same augmentation regimen as PointNet/PointNet++. The model is optimized with AdamW at a constant learning rate of \(10^{-4}\); loss weighting follows the effective-number scheme.

\subsection{Point Transformer v1 and v3}
\noindent For Point Transformer v1 \cite{zhao2021point} and v3 \cite{Wu_2024_pointtrasformerv3}, inputs are standardized to \(2{,}048\) points per item using farthest-point sampling. We apply the same normalization and geometric augmentations as above to avoid confounding from data processing. Both variants are trained for \(100\) epochs with AdamW (\(10^{-4}\) learning rate) and cross-entropy with class weighting as in Sec.~\ref{sec:experiment_setup}. Evaluation uses the identical point budget and normalization.

\subsection{OctFormer}
\noindent We follow the reference implementation of OctFormer \cite{wang2023octformer}. Coordinates are normalized to the unit sphere and encoded into an OCNN octree (depth \(8\), full\_depth \(2\)) with the associated neighborhood structures. The OctFormer backbone feeds an FPN-style head that upsamples multi-scale features and interpolates them to query points for per-point classification. Unless stated otherwise, we sample \(2{,}048\) points per item for parity with the other models and train with AdamW at \(10^{-4}\), cross-entropy (with \texttt{ignore\_label}, optional class weights and label smoothing), and the global settings of Sec.~\ref{sec:experiment_setup}.

\subsection{Superpoint Transformer}
\noindent We adopt the Superpoint Transformer (SPT) \cite{robert2023spt} in its reference implementation. 
As per the SPT methodology, each input scene is oversegmented into superpoints, so no further downsampling or chunking is required. 
Preprocessing and augmentation follow the DALES configuration in the official codebase. 
To remain faithful to the original setup, training uses stochastic gradient descent (SGD) with a learning rate of \(0.01\), weight decay of \(10^{-4}\), and a batch size of \(4\), which differs slightly from the global setting in Sec.~\ref{sec:experiment_setup}.

\subsection{Data Augmentation}
We acknowledge that the data augmentation is a common technique for improving model performance and increasing its generalization \cite{kong2023robo3d}.
However, we opted to omit data augmentation in our experimental setup, since it would have introduced an additional variable to the coherent domain gap analysis: 
The influence of data augmentation would make assessing the impact of real-to-synthetic data hardly tractable. 
%
\section{Additional Information about Dataset and Evaluation Result}
\label{sec:experiment_detail_result}

\begin{table*}[htb]
\centering
\scriptsize
\resizebox{\textwidth}{!}{%
\begin{tabular}{@{}lccccccccccc@{}}
\toprule
\multirow{2}{*}{Class / Metric} &
\multicolumn{5}{c}{PointNet} &
\multicolumn{5}{c}{PointNet++} \\
\cmidrule(lr){2-6}\cmidrule(lr){7-11}
 & 100S & 75S & 50S & 25S & 0S
 & 100S & 75S & 50S & 25S & 0S \\
\midrule
mIoU                  & 6.03 & 10.74 & 10.89 & 13.10 & \textbf{14.51}
                      & 9.72 & 20.95 & 23.18 & \textbf{25.38} & 23.39 \\
mAcc                  & 14.58 & 21.63 & 21.82 & 25.35 & \textbf{25.98}
                     & 19.75 & 29.37 & 35.22 & 30.84 & \textbf{38.03} \\
OA                    & 30.36 & 48.10 & 49.29 & 47.99 & \textbf{49.82}
                     & 34.39 & 62.80 & \textbf{65.36} & 63.27 & 63.15 \\
\midrule
RoadSurface           &  7.84 & 51.92 & 54.04 & 49.78 & \textbf{67.55}
                     & 60.90 & 73.90 & \textbf{81.20} & 80.30 & 72.50 \\
GroundSurface         & 16.45 & 24.11 & 17.27 & 21.90 & \textbf{28.55}
                     & 33.20 & 45.30 & \textbf{50.40} & 49.60 & 32.40 \\
CityFurniture         &  0.62 &  2.68 &  4.12 & \textbf{10.80} &  8.56
                     & 15.20 & 14.60 & 21.30 & 25.50 & \textbf{46.20} \\
Vehicle               &  0.00 &  0.00 &  0.00 &  0.00 &  0.00
                     & \textbf{3.50} &  0.00 &  0.18 &  0.51 &  0.01 \\
Pedestrian            &  0.00 &  0.00 &  0.00 &  0.00 &  0.00
                     & 0.00 &  0.00 &  0.00 &  \textbf{0.32} &  0.04 \\
WallSurface           & 39.97 & 37.01 & \textbf{45.28} & 43.07 & 30.87
                     & 0.20 & 56.90 & 63.50 & \textbf{65.70} & 63.20 \\
RoofSurface           &  0.00 &  0.06 &  \textbf{0.19} &  0.04 &  0.00
                     & 0.00 &  \textbf{0.80} &  0.30 &  0.10 &  0.30 \\
Door                  &  0.02 &  0.00 &  0.00 &  \textbf{0.11} &  0.00
                     & 0.00 &  0.00 &  0.20 &  0.10 &  \textbf{0.70} \\
Window                &  2.57 &  6.00 &  1.05 &  5.38 &  \textbf{6.39}
                     & 3.61 &  7.30 &  4.50 &  4.20 & \textbf{14.60} \\
BuildingInstallation  &  0.01 &  0.26 &  0.01 &  \textbf{4.53} &  0.15
                     & 0.00 &  0.70 &  3.00 &  1.70 &  \textbf{4.60} \\
SolitaryVegetationObject    &  4.24 &  1.04 &  5.78 & 11.40 & \textbf{14.79}
                     & 0.00 & \textbf{50.60} & 39.30 & 45.60 & 14.40 \\
Noise                 &  0.63 &  5.82 &  2.93 & 10.21 & \textbf{17.26}
                     & 0.00 &  1.30 & 14.30 & 30.90 & \textbf{31.70} \\
\bottomrule
\end{tabular}%
}
\caption{Per-class IoU (↑) under different S--R dataset mixtures. Models: \textbf{PointNet}~\cite{qi2017pointnet} and \textbf{PointNet++}~\cite{qi2017pointnet++}. Bold marks the best value for each model across mixtures.}
\label{tab:pair_pointnet_pointnetpp}
\end{table*}

\begin{table*}[htb]
\centering
\scriptsize
\resizebox{\textwidth}{!}{%
\begin{tabular}{@{}lccccccccccc@{}}
\toprule
\multirow{2}{*}{Class / Metric} &
\multicolumn{5}{c}{RandLA-Net} &
\multicolumn{5}{c}{KPConv} \\
\cmidrule(lr){2-6}\cmidrule(lr){7-11}
 & 100S & 75S & 50S & 25S & 0S
 & 100S & 75S & 50S & 25S & 0S \\
\midrule
mIoU                  & 8.98 & 13.25 & 15.73 & 16.89 & \textbf{17.71}  & 15.84 & 21.55 & 28.50 & 22.33 & \textbf{29.90} \\
mAcc                  & 25.62 & \textbf{31.27} & 28.48 & 29.82 & 30.97  & 38.12 & 33.59 & 37.79 & \textbf{42.91} & 42.01 \\
OA                    & 35.40 & 50.32 & \textbf{59.37} & 57.09 & 54.57  & 50.07 & 62.08 & 61.62 & 61.92 & \textbf{62.80} \\
\midrule
RoadSurface           &  3.12 & 60.53 & \textbf{70.35} & 62.01 & 60.26  & 30.64 & \textbf{69.26} & 54.26 & 56.31 & 54.59 \\
GroundSurface         & 23.31 & \textbf{29.51} & 29.41 & 25.27 & 26.65  & 24.67 & \textbf{36.49} & 31.91 & 33.20 & 35.96 \\
CityFurniture         &  2.33 & \textbf{15.91} &  9.99 & 14.38 & 12.93  & \textbf{18.25} & 10.09 & 12.67 &  1.85 & 14.57 \\
Vehicle  &  0.00 &  0.03 &  1.03 &  0.59 &  \textbf{3.27}  &  0.00 &  0.00 & 38.93 & \textbf{70.33} & 63.64 \\
Pedestrian & 0.00 & 0.00 & 0.00 & 0.00 & 0.00  & 0.00 & 0.00 & 0.00 & 0.00 & 0.00 \\
WallSurface           & 45.19 & 41.29 & 51.15 & \textbf{54.18} & 53.23  & 59.97 & 63.21 & 68.54 & \textbf{73.55} & 71.03 \\
RoofSurface           &  \textbf{0.32} &  0.17 &  0.29 &  0.19 &  0.18  &  0.82 &  0.01 &  0.47 &  0.00 &  \textbf{1.79} \\
Door                  & \textbf{18.22} &  2.51 &  1.41 &  0.16 &  3.25  &  0.00 &  0.00 &  0.00 &  0.00 &  0.00 \\
Window                &  3.82 &  2.83 &  9.51 &  8.75 &  \textbf{9.72}  &  1.86 &  0.43 & \textbf{18.21} &  1.17 &  2.89 \\
BuildingInstallation  &  0.94 &  1.68 &  2.91 &  \textbf{8.46} &  2.90  &  \textbf{7.85} &  3.02 &  0.85 &  0.85 &  3.72 \\
SolitaryVegetationObject    & 10.42 &  1.91 &  2.07 & 10.68 & \textbf{14.86}  & 45.74 & 70.51 & \textbf{81.45} &  0.41 & 77.13 \\
Noise                 &  0.10 &  2.64 & 10.63 & 17.95 & \textbf{25.27}  &  0.31 &  5.58 & \textbf{34.73} & 30.26 & 33.43 \\
\bottomrule
\end{tabular}%
}
\caption{Per-class IoU (↑) under different S--R dataset mixtures. Models: \textbf{RandLA-Net}~\cite{hu2020randlanet} and \textbf{KPConv}~\cite{thomas2019kpconv}. Bold marks the best value for each model across mixtures.}
\label{tab:pair_randla_kpconv}
\end{table*}

\begin{table*}[t]
\centering
\scriptsize
\resizebox{\textwidth}{!}{%
\begin{tabular}{@{}lccccccccccc@{}}
\toprule
\multirow{2}{*}{Class / Metric} &
\multicolumn{5}{c}{Point Transformer v1} &
\multicolumn{5}{c}{Point Transformer v3} \\
\cmidrule(lr){2-6}\cmidrule(lr){7-11}
 & 100S & 75S & 50S & 25S & 0S
 & 100S & 75S & 50S & 25S & 0S \\
\midrule
mIoU                  & 16.30 & 19.79 & 23.43 & 24.66 & \textbf{28.89}  & 14.13 & 19.29 & \textbf{25.30} & 24.64 & 25.24 \\
mAcc                  & 26.09 & 34.39 & 37.01 & 35.52 & \textbf{39.33}  & 28.07 & 34.09 & 43.16 & 40.33 & \textbf{44.09} \\
OA                    & 57.54 & 60.29 & 67.54 & \textbf{68.70} & 67.98  & 53.15 & 60.22 & \textbf{65.94} & 65.72 & 60.75 \\
\midrule
RoadSurface           & 62.41 & 61.89 & \textbf{80.33} & 77.93 & 70.44  & 64.42 & 68.85 & 70.74 & \textbf{71.17} & 42.36 \\
GroundSurface         & 31.94 & 37.54 & \textbf{45.54} & 37.09 & 41.67  & 32.53 & 32.66 & \textbf{33.27} & 32.80 & 27.53 \\
CityFurniture         & 25.89 & 19.50 & 13.51 & 14.23 & \textbf{44.83}  &  5.35 & 34.86 & 24.11 & 17.34 & \textbf{35.13} \\
Vehicle  &  0.00 & 0.51 & 4.41 & 1.51 & \textbf{6.54} &  0.00 &  0.00 &  0.00 &  0.00 &  0.00 \\
Pedestrian & 0.00 & 0.00 & \textbf{0.05} & 0.00 & 0.00  &  0.00 &  0.00 &  0.00 &  0.00 &  0.00 \\
WallSurface           & 53.15 & 61.42 & 64.39 & 65.32 & \textbf{67.10}  & 48.15 & 54.79 & 63.42 & 64.63 & \textbf{69.69} \\
RoofSurface           &  0.10 &  0.29 &  \textbf{0.92} &  0.30 &  0.08  &  0.01 &  0.10 &  0.10 &  0.00 & \textbf{0.36} \\
Door                  &  0.01 &  0.13 &  0.51 &  \textbf{0.66} &  0.01  &  0.00 &  0.81 & \textbf{2.77} &  0.08 &  1.62 \\
Window                &  3.28 &  5.61 &  7.61 & 11.78 & \textbf{31.29}  &  3.09 &  5.17 &  4.92 &  4.35 & \textbf{10.70} \\
BuildingInstallation  &  3.08 &  5.24 &  6.69 &  \textbf{8.87} &  3.41  &  8.40 &  2.53 & \textbf{9.95} &  9.30 &  4.44 \\
SolitaryVegetationObject    & 15.54 & 35.35 & 34.70 & 43.22 & \textbf{52.31}  &  6.72 & 12.90 & 60.86 & 60.80 & \textbf{70.39} \\
Noise                 &  0.16 &  9.98 & 22.43 & \textbf{35.04} & 28.96  &  0.84 & 18.80 & 33.50 & 35.19 & \textbf{40.69} \\
\bottomrule
\end{tabular}%
}
\caption{Per-class IoU (↑) under different S--R dataset mixtures. Models: \textbf{Point Transformer v1}~\cite{zhao2021point} and \textbf{Point Transformer v3}~\cite{Wu_2024_pointtrasformerv3}. Bold marks the best value for each model across mixtures.}
\label{tab:pair_ptv1_ptv3}
\end{table*}

\begin{table*}[htb]
\centering
\scriptsize
\resizebox{\textwidth}{!}{%
\begin{tabular}{@{}lccccccccccc@{}}
\toprule
\multirow{2}{*}{Class / Metric} &
\multicolumn{5}{c}{OctFormer} &
\multicolumn{5}{c}{Superpoint Transformer} \\
\cmidrule(lr){2-6}\cmidrule(lr){7-11}
 & 100S & 75S & 50S & 25S & 0S
 & 100S & 75S & 50S & 25S & 0S \\
\midrule
mIoU                  & 13.07 & 14.17 & 14.22 & 13.91 & \textbf{17.65}  & 14.31 & 17.01 & 14.22 & \textbf19.61{} & 15.96 \\
mAcc                  & 22.84 & 23.99 & 26.28 & 26.15 & \textbf{27.19}  & 24.28 & 29.40 & 26.42 & \textbf{31.79} & 28.29 \\
OA                    & 53.30 & 55.34 & 49.71 & 50.97 & \textbf{56.28}  & 54.17 & \textbf{58.62} & 54.63  & 56.98 & 54.64 \\
\midrule
RoadSurface           & 49.32 & 51.58 & 35.34 & 54.81 & \textbf{71.98}  & \textbf{77.74} & 73.73 & 61.19 & 63.51 & 53.16 \\
GroundSurface         & 16.01 & 16.34 & \textbf{22.60} & 19.60 & 16.99  & 18.09 & 30.35 & 16.83 & \textbf{31.35} & 25.92 \\
CityFurniture         &  0.92 &  2.18 & \textbf{15.24} &  0.01 &  0.14  &  1.22 &  \textbf{3.70} &  0.85 &  2.49 &  0.47 \\
Vehicle  &  0.00 &  0.00 &  0.00 &  0.07 & \textbf{20.82}  &  0.00 &  0.26 & 0.55 &  \textbf{4.82} &  0.22  \\
Pedestrian & 0.00 & 0.00 & 0.00 & 0.00 & 0.00    &  0.00 &  0.00 & 0.00 &  0.00 &  0.00  \\
WallSurface           & 56.22 & \textbf{62.14} & 60.36 & 45.79 & 38.96  & 38.97 & 47.32 & 55.57 & 56.98 & \textbf{63.19} \\
RoofSurface           &  0.00 &  0.00 &  0.00 &  0.00 &  0.00  &  2.11 &  2.60 &  2.84 &  \textbf{3.21} &  1.25 \\
Door                  &  0.00 &  0.00 &  0.00 &  0.00 &  0.00  &  \textbf{0.30} &  0.09 &  0.00 &  0.00 &  0.00 \\
Window                &  0.00 &  0.00 &  \textbf{0.76} &  0.00 &  0.00  &  \textbf{8.94} &  3.89 &  1.43 &  2.99 &  0.55 \\
BuildingInstallation  &  0.00 &  0.00 &  \textbf{4.02} &  0.00 &  0.00  &  2.29 &  0.75 &  0.01 &  1.30 &  \textbf{2.63} \\
SolitaryVegetationObject    & 34.33 & 37.85 & 32.25 & 26.72 & \textbf{47.30}  & 21.54 & 34.45 & 16.70 & \textbf{48.13} & 21.12 \\
Noise                 &  0.00 &  0.00 &  0.08 & \textbf{19.93} & 15.57  &  0.46 & 6.96 & 14.70 & 20.50 & \textbf{23.07}  \\
\bottomrule
\end{tabular}%
}
\caption{Per-class IoU (↑) under different S--R dataset mixtures. Models: \textbf{OctFormer}~\cite{wang2023octformer} and \textbf{Superpoint Transformer}~\cite{robert2023spt}. Bold marks the best value for each model across mixtures.}
\label{tab:pair_octformer_superpoint_vehicle_ped}
\end{table*}



\begin{figure}[htb]
    \centering
    \includegraphics[width=\columnwidth]{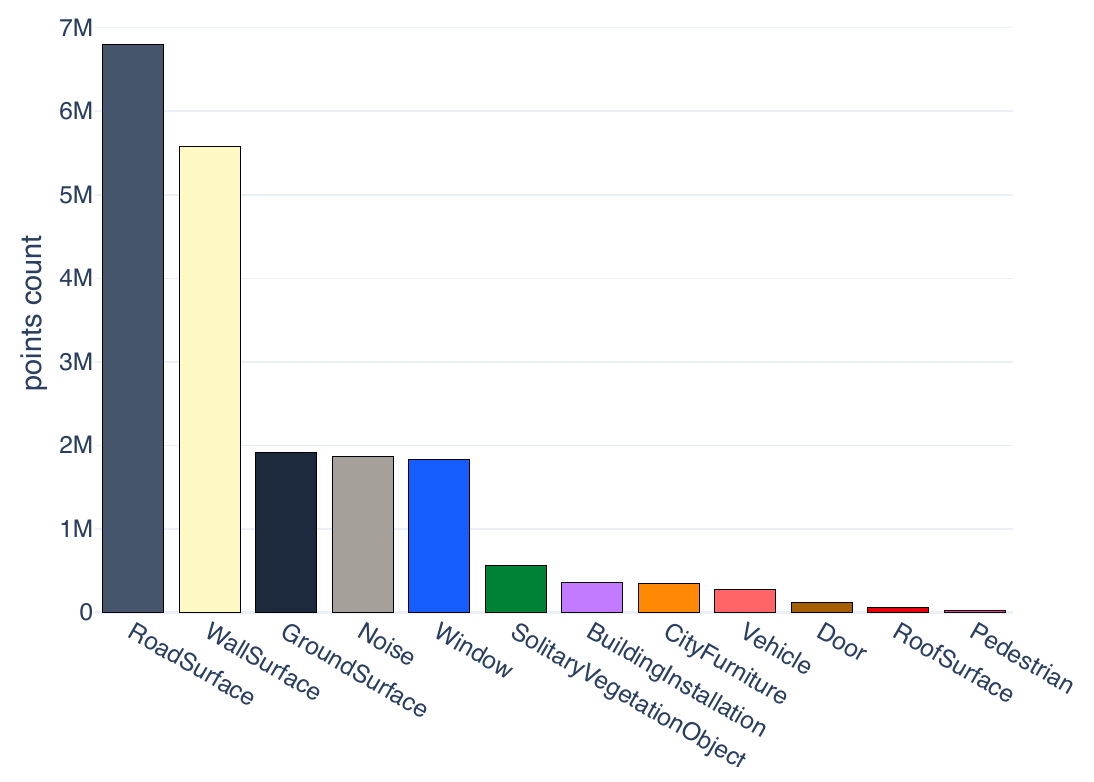}
    \caption{The class distribution of the real-only test set.}
    \label{fig:supp_test_100.pdf}
\end{figure}
\begin{figure}[htb]
    \centering
    \includegraphics[width=\columnwidth]{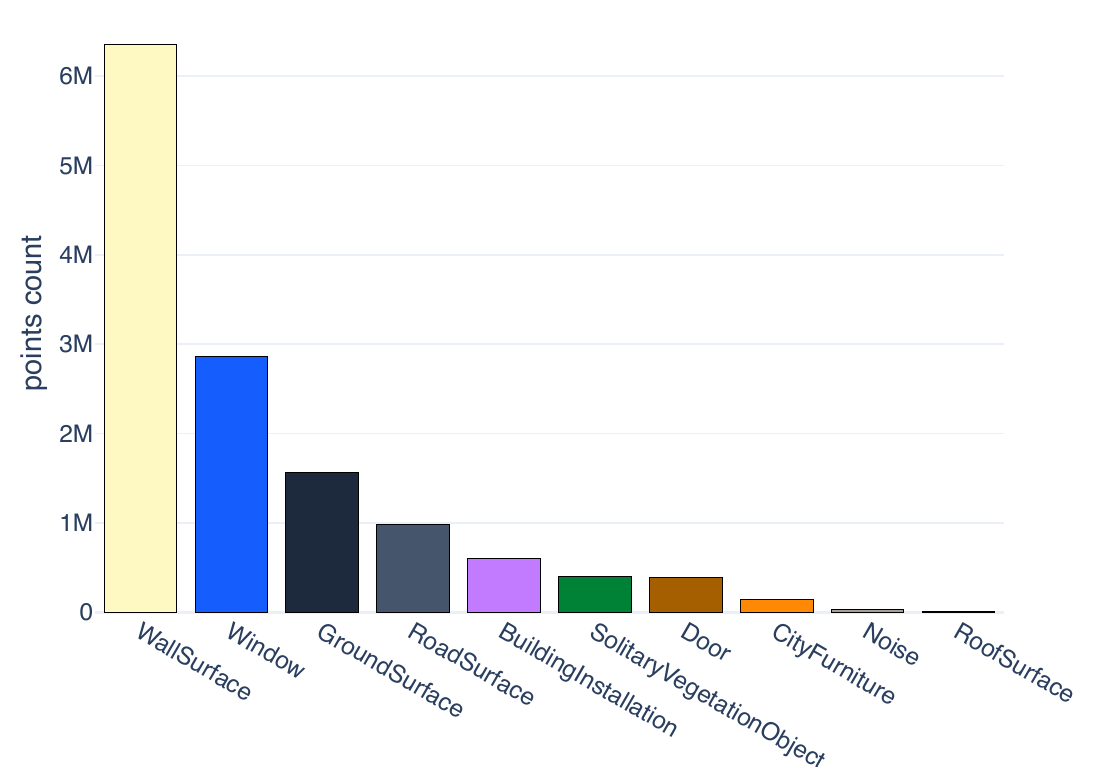}
    \caption{The class distribution of the synthetic-only test set, which is not used in training and evaluation.}
    \label{fig:supp_test_0.pdf}
\end{figure}

\begin{figure}[htb]
    \centering
    \includegraphics[width=\columnwidth]{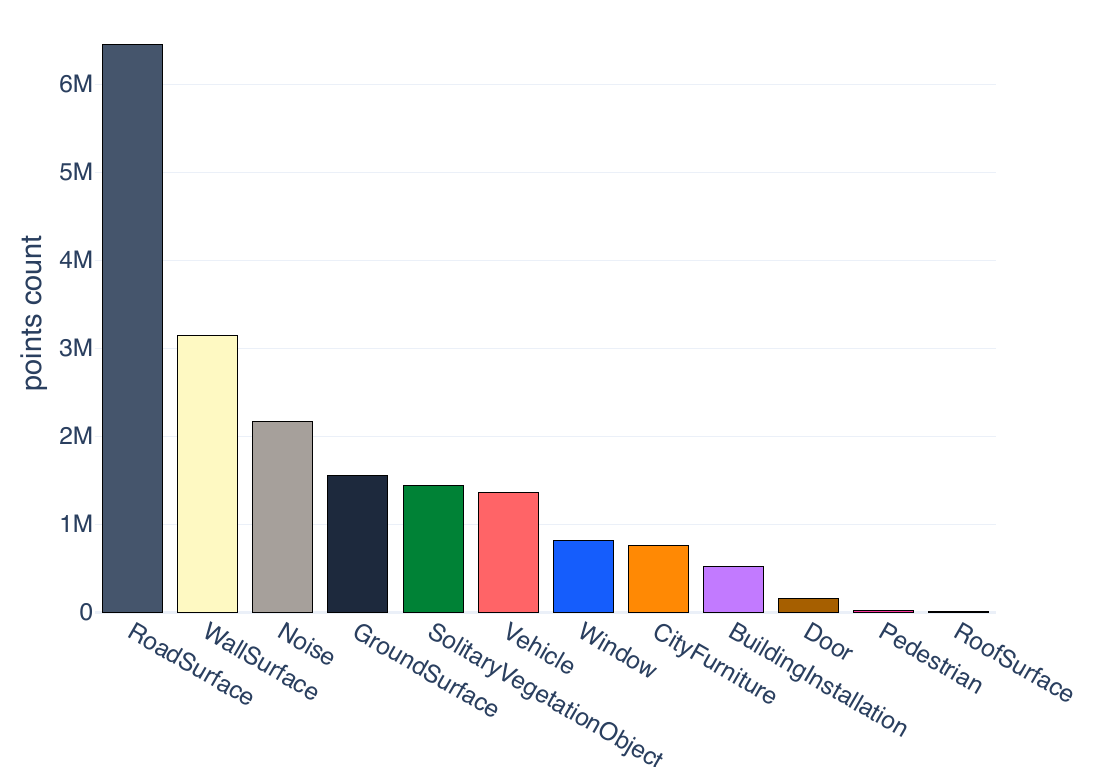}
    \caption{The class distribution of the real-only validation set.}
    \label{fig:supp_val_100}
\end{figure}
\begin{figure}[t]
    \centering
    \includegraphics[width=\columnwidth]{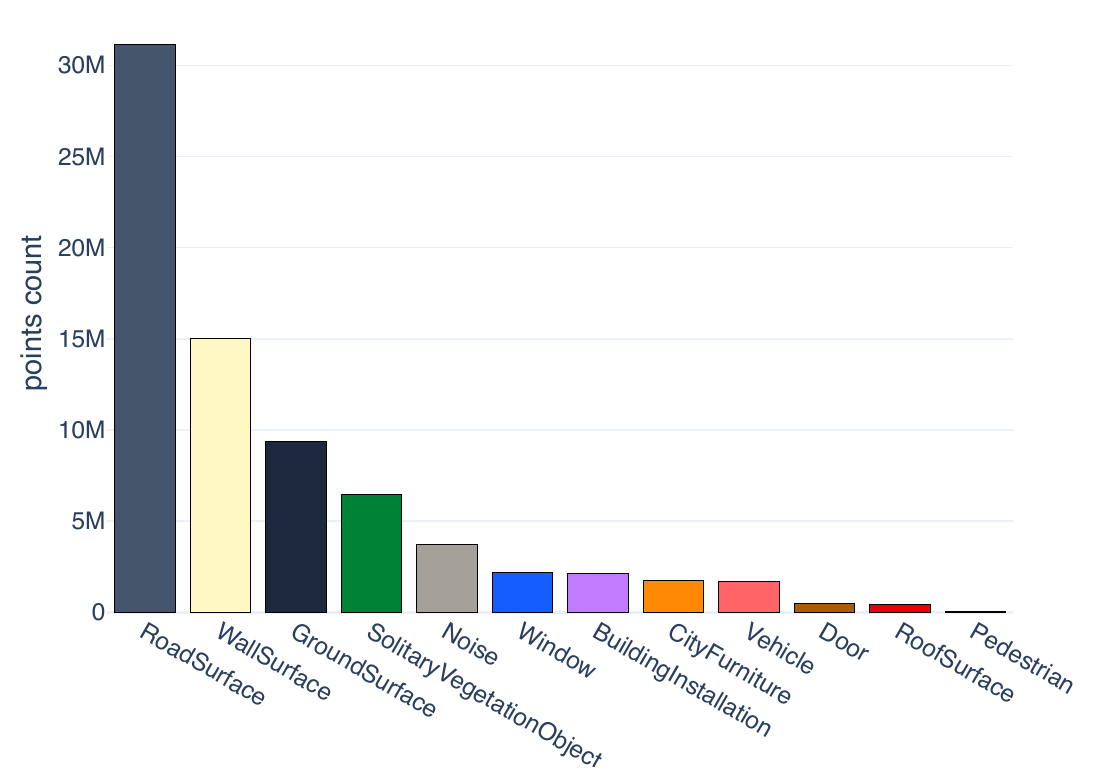}
    \caption{The class distribution of the 0S--100R train set.}
    \label{fig:supp_train_100}
\end{figure}
\begin{figure}[t]
    \centering
    \includegraphics[width=\columnwidth]{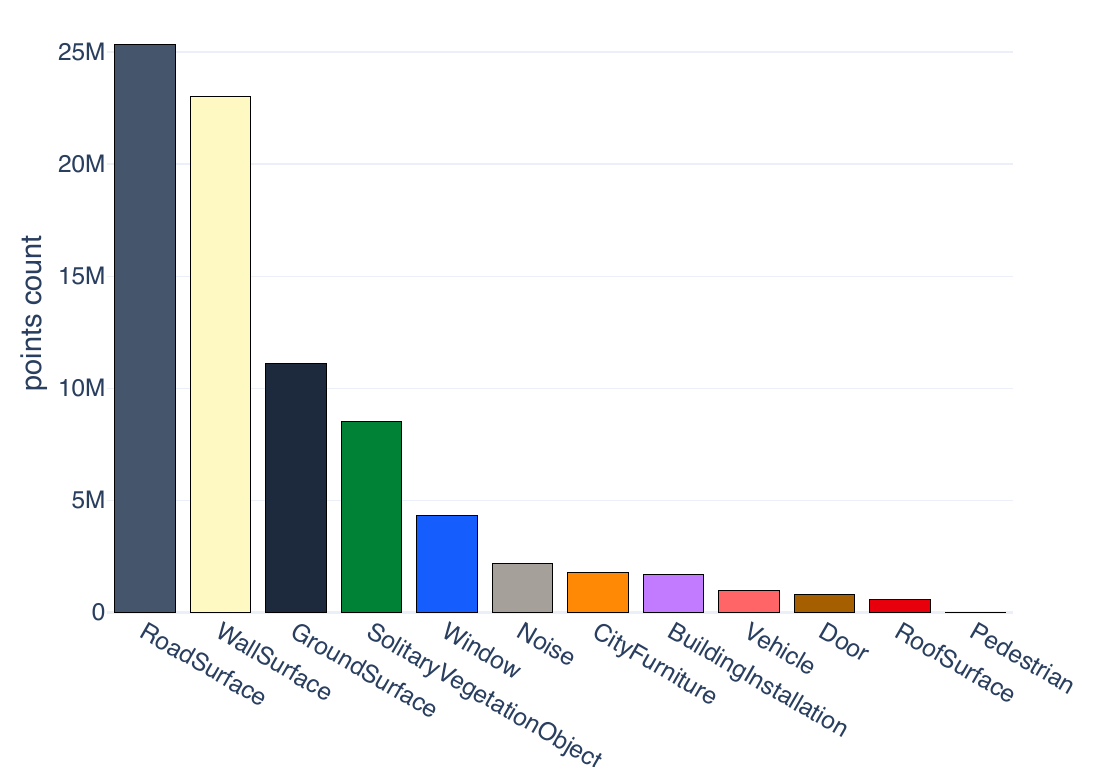}
    \caption{The class distribution of the 25S--75R train set.}
    \label{fig:supp_train_75}
\end{figure}
\begin{figure}[t]
    \centering
    \includegraphics[width=\columnwidth]{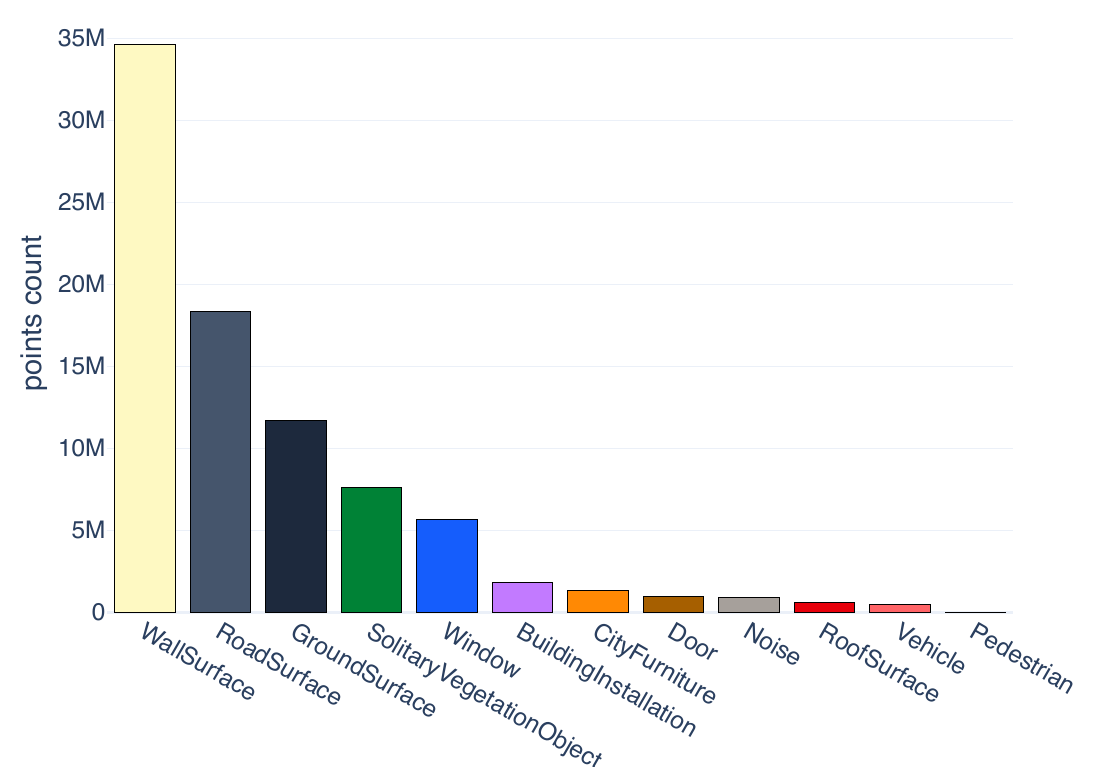}
    \caption{The class distribution of the 50S--50R train set.}
    \label{fig:supp_train_50}
\end{figure}
\begin{figure}[t]
    \centering
    \includegraphics[width=\columnwidth]{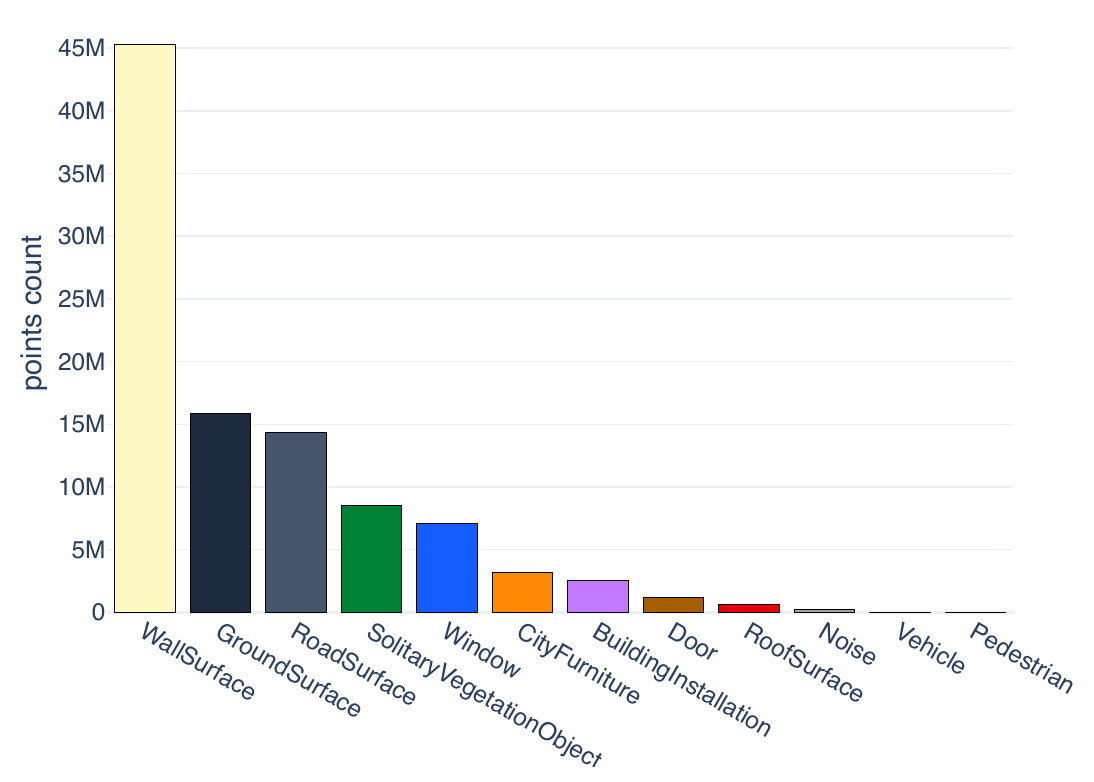}
    \caption{The class distribution of the 75S--25R train set.}
    \label{fig:supp_train_25}
\end{figure}
\begin{figure}[t]
    \centering
    \includegraphics[width=\columnwidth]{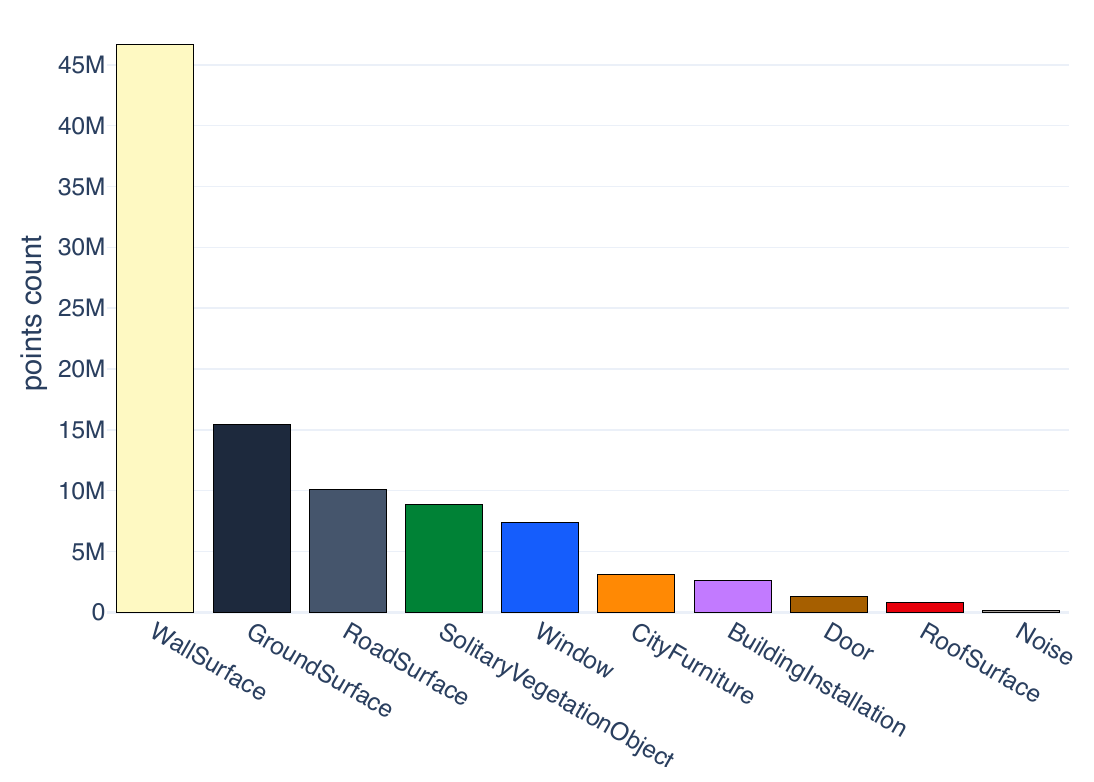}
    \caption{The class distribution of the 100S--0R train set.}
    \label{fig:supp_train_0}
\end{figure}

\end{document}